\documentclass[lettersize,journal]{IEEEtran}
\usepackage{amsmath,amsfonts}
\usepackage{algorithmic}
\usepackage{algorithm}
\usepackage{array}
\usepackage[caption=false,font=normalsize,labelfont=sf,textfont=sf]{subfig}
\usepackage{textcomp}
\usepackage{stfloats}
\usepackage{url}
\usepackage{verbatim}
\usepackage{graphicx}
\usepackage{cite}
\usepackage{bm}
\usepackage{booktabs}
\usepackage{xcolor}
\usepackage{subfig}
\begin{document}

\title{Bayesian-based Online Label Shift Estimation with Dynamic Dirichlet Priors}

\author{Jiawei Hu,~\IEEEmembership{Student Member,~IEEE,}
        Javier A. Barria,~\IEEEmembership{Member,~IEEE}
        % <-this % stops a space
\thanks{The authors are with the Department of Electrical and Electronic Engineering, Imperial College London, London SW7 2AZ, U.K. (e-mail: j.hu24@imperial.ac.uk; j.barria@imperial.ac.uk).}% <-this % stops a space
% \thanks{Manuscript received April 19, 2021; revised August 16, 2021.}
}

% The paper headers
% \markboth{Journal of \LaTeX\ Class Files,~Vol.~14, No.~8, August~2021}%
% {Shell \MakeLowercase{\textit{et al.}}: A Sample Article Using IEEEtran.cls for IEEE Journals}

% \IEEEpubid{0000--0000/00\$00.00~\copyright~2021 IEEE}
% Remember, if you use this you must call \IEEEpubidadjcol in the second
% column for its text to clear the IEEEpubid mark.

\maketitle

\begin{abstract}
Label shift, a prevalent challenge in supervised learning, arises when the class prior distribution of test data differs from that of training data, leading to significant degradation in classifier performance. To accurately estimate the test priors and enhance classification accuracy, we propose a Bayesian framework for label shift estimation, termed Full Maximum A Posterior Label Shift (FMAPLS), along with its online version, online-FMAPLS. Leveraging batch and online Expectation–Maximization (EM) algorithms, these methods jointly and dynamically optimize Dirichlet hyperparameters $\boldsymbol{\alpha}$ and class priors $\boldsymbol{\pi}$, thereby overcoming the rigid constraints of the existing Maximum A Posterior Label Shift (MAPLS) approach. Moreover, we introduce a linear surrogate function (LSF) to replace gradient-based hyperparameter updates, yielding closed-form solutions that reduce computational complexity while retaining asymptotic equivalence. The online variant substitutes the batch E-step with a stochastic approximation, enabling real-time adaptation to streaming data. Furthermore, our theoretical analysis reveals a fundamental trade-off between online convergence rate and estimation accuracy. Extensive experiments on CIFAR100 and ImageNet datasets under shuffled long-tail and Dirichlet test priors demonstrate that FMAPLS and online-FMAPLS respectively achieve up to 40\% and 12\% lower KL divergence and substantial improvements in post-shift accuracy over state-of-the-art baselines, particularly under severe class imbalance and distributional uncertainty. These results confirm the robustness, scalability, and suitability of the proposed methods for large-scale and dynamic learning scenarios.
\end{abstract}


\section{Introduction} \label{Section1}
\subsection{Context and Scope} \label{Section1-1}
\IEEEPARstart{D}{ue} to the inherent complexity and heterogeneity of large-scale datasets, the statistical properties of training datasets in supervised learning tasks often fail to accurately represent those of the test datasets encountered in practice, thereby violating the common assumption that both training and test data are drawn from the same underlying distribution in real-world applications \cite{Storkey2009}. This discrepancy in data distributions, commonly referred to as \textit{distribution shift} \cite{Kulinski2023}, has been widely recognized as one of the main factors leading to performance degradation of supervised classification models once they are deployed beyond the training environment. Specifically, when the data distribution encountered during test inference differs from that of the training phase, the learned decision boundaries may no longer align well with the characteristics of the unseen data, thereby reducing the generalization ability of the model.

As a common type of distribution shift, \textit{label shift} occurs in scenarios where the prior distribution of the class labels $P(Y)$ changes between training and test datasets, while the likelihood distribution of the features given the labels, $P(X|Y)$, remains unchanged \cite{Moreno-Torres2012}. This phenomenon is particularly relevant in applications such as medical diagnosis \cite{Ma2022}, spam email detection \cite{Martino2023}, and text topic classification \cite{Costache2025}, where the prevalence of certain classes may vary significantly over time or across environments, making it crucial to develop strategies that can explicitly account for label distribution mismatches.

A straightforward strategy to address label shift is to retrain separate classification models for different data distributions. However, retraining demands substantial computational resources and time \cite{Mahadevan2023}, making it impractical in many real-world applications. Therefore, it is necessary to develop approaches that mitigate the adverse effects of label shift without modifying the trained model. Leveraging Bayes’ theorem \cite{Murphy2012, Bishop2006}, the test prior can be employed to reweight the soft posterior probabilities produced by the trained classification model, thereby improving predictive performance on unseen test data while preserving the model architecture. This correction can enhance model adaptability under label shift, highlighting the critical importance of accurately estimating the test dataset prior probability.

With the rapid development of Internet of Things (IoT), terminal devices are increasingly deployed in various application scenarios, yet they typically operate under stringent constraints in terms of computational capacity, memory, and energy consumption \cite{Xie2021, Olfat2017}. When such devices are required to handle large-scale datasets or continuous data streams, traditional batch-based offline algorithms become infeasible, as they demand substantial resources and are unable to meet real-time processing requirements \cite{Bottou1998, Gama2014}. To address this limitation, it is essential to design lightweight and efficient online schemes for prior probability estimation that can operate effectively on streaming data. Such approaches not only reduce computational and storage overhead but also ensure adaptive correction of label shift in dynamic environments, thereby improving robustness and reliability of learning systems.

\begin{figure*}[t]
    \centering
    \includegraphics[width=6.2in]{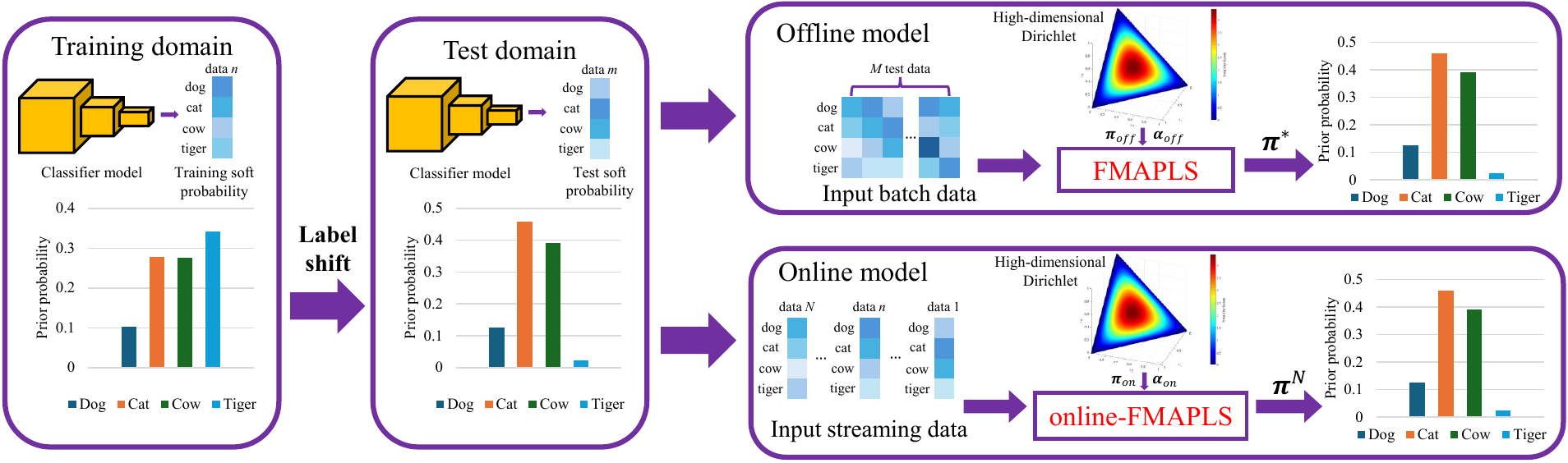} % 设置图片宽度为整页宽度
    \caption{System architecture of the proposed Bayesian-based label shift estimation framework. The offline algorithm (FMAPLS) and its online variant (online-FMAPLS) estimate the class prior distribution of the test domain using high-dimensional Dirichlet modeling and EM-based inference, thus supporting both batch and streaming data processing.}
    \label{Figure1}
    \vspace{-16pt}
\end{figure*}

As illustrated in Fig.~\ref{Figure1}, this paper introduces a Bayesian-based label shift estimation framework, called Full Maximum A Posterior Label Shift (FMAPLS), together with its online variant, online-FMAPLS. The proposed approach employs a high-dimensional Dirichlet distribution to model test dataset class priors, thereby offering a probabilistic and flexible representation for prior estimation. To accurately and iteratively estimate test priors, the framework employs the Expectation–Maximization (EM) algorithm in offline settings and its online counterpart (online-EM) for streaming scenarios, thereby recalibrating soft probabilities produced by the classifier and enhancing its performance under label shift conditions.

\vspace{-20pt}
\subsection{Related Work} \label{Section1-2}
The EM algorithm was one of the earliest approaches to address label shift, which updates the test priors iteratively based on classifier outputs \cite{Saerens2002, Chan2005}. To address the above EM algorithm's explicit requirement for training likelihood probability problem, \cite{Zhang2013} utilized the kernel mean embedding of the marginal and conditional distributions. The Black Box Shift Estimation (BBSE) method later leverages the confusion matrix of a held-out training set to identify and correct distribution shifts, outperforming previous kernel-based approaches that scale poorly with dataset size and struggle on high-dimensional data \cite{Lipton2018}. Building upon BBSE, Regularized Learning of Label Shift (RLLS) incorporated a regularization framework to improve importance weight estimation between labeled training and unlabeled test data \cite{Azizzadenesheli2019}, and empirical studies showed that RLLS outperforms BBSE in both weight estimation precision and classification performance. Zhao \textit{et al.} \cite{Zhao2021} further proposed Mediated Active Learning under Label Shift (MALLS) algorithm for label shift active learning. More recently, EM-based methods have been revisited and demonstrated superior performance in limited-data scenarios, outperforming both BBSE and RLLS on benchmark datasets such as CIFAR10 and MNIST \cite{Alexandari2020, Garg2020}. Despite these advances, existing techniques have primarily been evaluated on relatively small datasets and rarely account for the influence of class-imbalanced training data in label shift estimation.

Ye \textit{et al.} \cite{Ye2024} recently proposed a Bayesian framework for label shift estimation that models class priors with a high-dimensional Dirichlet distribution and implements Maximum A Posterior Label Shift (MAPLS) via EM-based offline optimization. While MAPLS outperforms BBSE and RLLS on large-scale datasets such as CIFAR100, ImageNet, and their long-tail variants, its Adaptive Prior Learning (APL) relies on two rigid assumptions: uniform Dirichlet hyperparameters $\bm{\alpha}$ in each EM iteration and empirically defined update equations, limiting adaptability to complex or rapidly changing distributions \cite{Zhang2024, Meng2025}. Building on MAPLS, our previous work \cite{Hu2025} introduced FMAPLS, which dynamically adapts hyperparameters and jointly optimizes class priors through EM iterations. Numerical experiments demonstrated that FMAPLS achieves superior effectiveness and robustness, particularly under highly imbalanced and highly uncertain prior scenarios.

While the above approaches are able to mitigate label shift to some extent, they all rely on batch-based algorithm, which requires substantial computational resources and time, thereby restricting their applicability to offline scenarios \cite{Shen2017}. With the increasing miniaturization of terminal devices and the growing prevalence of streaming data, there is a pressing need for online label shift estimation methods that can process data sequentially and adapt model parameters immediately after the arrival of each new sample \cite{Jenkins2025, Chen2012}. In 2021, Wu \textit{et al.} \cite{Wu2021} were the first to address online label shift estimation by employing online gradient descent (OGD) and Follow the History (FTH) method. Building on this foundation, Bai \textit{et al.} \cite{Bai2022} introduced a reconstructed unbiased risk estimator and provided provable theoretical guarantees for online label shift estimation, thereby strengthening the theoretical underpinnings of this line of research. More recently, Baby \textit{et al.} \cite{Baby2023} advanced this direction by reducing the adaptation problem to online regression and developing a novel algorithm that guarantee optimal dynamic regret, while simultaneously achieving both theoretical optimality and strong empirical performance. However, existing online methods do not address label shift estimation from a Bayesian perspective and, like offline solutions, fail to account for the impacts of large-scale and class-imbalanced training data.

\vspace{-10pt}
\subsection{Main Contributions} \label{Section1-3}
In this paper, we propose a Bayesian-based online label shift estimation algorithm, termed online-FMAPLS, developed within the online EM framework. In the online-FMAPLS algorithm, the E-step of the conventional EM procedure is replaced with a stochastic approximation step, allowing adaptation to streaming data while reducing computational complexity \cite{Wang2023, He2023}. Building upon this formulation, we further derive the Cramér–Rao bound (CRB) and analyze the convergence rate of the proposed online method. Finally, extensive experiments on large-scale CIFAR100 and ImageNet datasets are conducted to validate its estimation performance for test priors. The main contributions are summarized as follows:

\textbf{Dynamic Dirichlet Hyperparameter Adaptation}: We provide a theoretical characterization of the estimation performance for Bayesian label shift by deriving the CRB, alongside a novel estimation framework. Our proposed EM-based method dynamically and jointly optimizes class priors $\bm{\pi}$ and Dirichlet hyperparameters $\bm{\alpha}$, relaxing the rigid constraints of conventional MAPLS algorithm. This adaptation significantly improves the model's flexibility and expressivity, enabling the effective capture of high imbalance and high uncertainty priors in test datasets \cite{Hu2025}.

\textbf{Online Variant for Streaming Data}: We develop an online version, termed online-FMAPLS, which replaces the traditional E-step with a stochastic approximation step, thereby reducing the computational complexity from the batch-dependent $\mathcal{O}(NK)$ to $\mathcal{O}(K)$ per iteration. This design enables sequential parameter updates in streaming environments and makes the method highly suitable for large-scale and real-time applications. Furthermore, we analyze the convergence behavior of the proposed online-FMAPLS algorithm and reveal a fundamental trade-off between convergence rate and estimation precision.

\textbf{Surrogate-Based Closed-Form Hyperparameter Updates}: To overcome the absence of closed-form analytical updates for the Dirichlet hyperparameter $\bm{\alpha}$, we introduce a linear surrogate function (LSF) that reduces the computational complexity from $\mathcal{O}(T_{\text{grad}}K)$ to $\mathcal{O}(K)$, thereby making the method suitable to large-scale settings. Beyond computational efficiency, the surrogate-based formulation removes the sensitivity of $\bm{\alpha}$ updates to initialization and learning-rate tuning, ensuring stable convergence in both batch and streaming scenarios. Moreover, our analytical results show that the LSF-based update is asymptotically equivalent to gradient ascent procedure \cite{Hu2025}, and the convergence–accuracy trade-off is governed by the LSF parameter $c$.

Compared to our prior work \cite{Hu2025}, this paper improves mainly in four aspects: (1) Providing a detailed derivation of FMAPLS algorithm and establishing its corresponding CRB, thereby strengthening the theoretical guarantees of the proposed estimation framework. (2) Moreover, \cite{Hu2025} focused exclusively on offline scenarios, this paper further develops an online EM–based approach for online label shift estimation, enabling adaptation to sequential data with reduced computational complexity. (3) We analysis the convergence rate of online-FMAPLS algorithm and show a fundamental trade-off between  convergence rate and estimation precision. (4) In addition, we conduct more comprehensive experiments on CIFAR100 and ImageNet datasets to evaluate both the offline and online algorithms, and perform ablation studies for a thorough comparative analysis.

\vspace{-10pt}
\subsection{Organization} \label{Section1-4}
The remainder of this paper is organized as follows. Section~\ref{Section2} formalizes the problem setting, defining label shift and the prior assumptions on training and test datasets. Section~\ref{Section3} introduces the batch EM-based FMAPLS algorithm, detailing its E-step and M-step updates. To handle the absence of a closed-form update for $\bm{\alpha}$, a surrogate replacement is proposed, and its convergence and asymptotic equivalence are analyzed. Section~\ref{Section4} develops an online variant, termed online-FMAPLS, which updates parameters in a online manner for large-scale or real-time applications. Section~\ref{Section5} compares computational complexity between offline and online algorithms as well as between gradient ascent and surrogate-based updates. Section~\ref{Section6} presents experimental results on CIFAR100 and ImageNet, demonstrating the proposed methods’ effectiveness and robustness under highly imbalanced and uncertain test priors. Finally, Section~\ref{Section7} concludes the paper.

\vspace{-10pt}
\section{Problem Definition and Setting} \label{Section2}
\subsection{Overall setting and label shift} \label{Section2-1}
Consider $\boldsymbol{X}$ denotes the input space and $\boldsymbol{Y}=\{1,2,\ldots,K\}$ the label space with $K$ classes. A neural network-based classifier is trained to learn a mapping $f:\boldsymbol{X}\rightarrow\boldsymbol{Y}$ that outputs, for each sample $X\in\boldsymbol{X}$, a soft posterior probability $\boldsymbol{P}(\boldsymbol{Y}|X)$, and the predicted label $\hat{Y}$ corresponds to its maximum value. This classifier models the relationship between features and labels under the training distribution. However, when the test distribution differs from the training one, the estimated posteriors may become biased, leading to degraded performance. Therefore, accurately estimating and correcting distribution shifts is essential for maintaining robust prediction across mismatched training and testing environments.

This work primarily focuses on the \textit{label shift} setting, where the prior distributions of the training and test domains differ, while the conditional likelihood distribution $P(\boldsymbol{X}|\boldsymbol{Y})$ is assumed to remain invariant:
\begin{align}
    \label{eq1}
    \left\{
        \begin{aligned}
        &P(\bm{X}_{s}|\bm{Y}_{s}) = P(\bm{X}_{t}|\bm{Y}_{t})\\
        &P(\bm{Y}_{s}) \ne P(\bm{Y}_{t})\\
    \end{aligned}
    \right.
\end{align}
where $\bm{X}_{s}$ and $\bm{Y}_{s}$ denote the training domain input data and labels, respectively, while $\bm{X}_{t}$ and $\bm{Y}_{t}$ represent the test domain counterparts. Such a shift alters the class proportions between training and test data without changing the feature–label relationship, thereby making it possible to adapt a pre-trained classifier by appropriately estimating test prior probabilities and reweighting the classifier's output.

\vspace{-10pt}
\subsection{Training and test prior setting} \label{Section2-2}
To simulate the discrepancy between prior distributions of training and test data and to study the impact of imbalanced data on label shift estimation, the classification model is trained on data whose class priors follow a standard long-tail distribution with an imbalance ratio $\rho$ \cite{Cao2019}, while the test data are constructed with priors following either a shuffled long-tail or a Dirichlet distribution \cite{Ye2024}. The class frequency vector of the training dataset $\bm{n_{s}}$ and test dataset $\bm{n_{t}}$ can be written as:
\begin{equation}
	\label{eq2}
	\bm{n}_{s} = f_{LT}(\bm{N_{s}}, \rho, K) = \lfloor \bm{N_{s}}  \rho^{\frac{i}{K}} \rfloor
\end{equation}
\vspace{-16pt}
\begin{align}
    \label{eq3}
    \bm{n}_{t}^{(l)} = \left\{
        \begin{aligned}
        & \sigma(f_{LT}(\bm{N_{t}}, \rho, K))\;\;\;\;l=\text{shuffle}\\
        & \lfloor \bm{N_{t}} \circ Dir(\bm{\alpha}) \rfloor\;\;\;\;\;\;\;\;l=\text{Dirichlet}
    \end{aligned}
    \right.
\end{align}
where $\bm{N_{s}}$ and $\bm{N_{t}}$ denote the class sizes in the training and test domains, respectively. The operator $\sigma(\cdot)$ indicates a shuffling function, and $Dir(\bm{\alpha})$ denotes a sampled Dirichlet prior probability vector parameterized by a $K$-dimensional hyperparameter vector $\bm{\alpha}$. The symbol $\circ$ refers to the Hadamard (element-wise) product, and $\lfloor\cdot\rfloor$ denotes the floor operation. Notably, the shuffled long-tail prior is employed to evaluate the algorithm’s robustness under severe class imbalance, whereas the Dirichlet prior is adopted to assess its adaptability in scenarios with high prior uncertainty.

\vspace{-8pt}
\subsection{Dirichlet distribution} \label{Section2-3}
The Dirichlet distribution $Dir(\bm{\pi}; \bm{\alpha})$ can be regarded as the multivariate extension of the beta distribution \cite{Pham-Gia2009}. It is parameterized by a $M$-dimensional hyperparameter $\bm{\alpha} \in \mathbb{R}^{M}{>0}$, which controls the distribution’s shape, and is supported on the $(M-1)$-dimensional standard simplex $\Delta^{M-1}=\{\bm{\pi}\in\mathbb{R}^{M}|\pi_i\geq0, \sum_{i=1}^{M}\pi_i=1\}$. The probability density function (PDF) of Dirichlet distribution can be written as:
\begin{equation}
	\label{eq4}
	P(\bm{\pi}; \bm{\alpha})=Dir(\bm{\pi}; \bm{\alpha})=\frac{1}{\boldsymbol{B}(\boldsymbol{\alpha})}\prod_{j=1}^{M}\pi_{j}^{\alpha_{j}-1}
\end{equation}
\vspace{-2pt}
\begin{equation}
	\label{eq5}
	\boldsymbol{B}(\boldsymbol{\alpha})=\frac{\prod_{i=1}^{M} \Gamma(\alpha_{i})}{\Gamma(\sum_{j=1}^{M}\alpha_{j})}
\end{equation}
where $\Gamma(\cdot)$ is the Gamma function. It is worth noting that when $\bm{\alpha}=\bm{1}$, the Dirichlet distribution degenerates to the uniform distribution. Due to its support on the probability simplex and its role as the conjugate prior for both categorical and multinomial distributions, the Dirichlet distribution is widely adopted in Bayesian statistics to model class priors \cite{Tahsin2024, Olkin2003, Steck2002, Xu2024}. Accordingly, this work employs the Dirichlet distribution and applies batch and online EM algorithms for offline and online estimation of test priors, respectively.

\vspace{-6pt}
\section{Batch EM Method} \label{Section3}
This section presents the batch EM-based test prior estimation algorithm, referred to as FMAPLS. The algorithm is capable of simultaneously utilizing $N$ soft probability outputs $\{\bm{P}(\bm{Y}|X_t)\}_{t=1}^{N}$ produced by the classifier.

\vspace{-8pt}
\subsection{FMAPLS algorithm} \label{Section3-1}
Assume the neural network-based classifier has $N$ input data $\boldsymbol{X}=\{x_i, x_2, ..., x_N\}$, the complete posterior probability of the Dirichlet hyperparameter and the test prior probability set $\boldsymbol{\theta}:\{\boldsymbol{\alpha};\boldsymbol{\pi}\}$ can be written as \cite{Hu2025}:
\begin{align}
    \label{eq6}
    P(\boldsymbol{\theta}|\boldsymbol{X}, \boldsymbol{Y})=\frac{1}{Z}P(\boldsymbol{\theta})\prod_{i=1}^{N}P(X_s=x_i|Y_s=y_i)P(Y_t=y_i|\boldsymbol{\theta})
\end{align}
where $Z$ is a constant and $P(\boldsymbol{\theta})$ is the Dirichlet PDF $Dir(\bm{\pi}; \bm{\alpha})$ given in (\ref{eq4}) and (\ref{eq5}). Based on this, the batch EM is applied to iteratively estimate the parameter set $\bm{\theta}$.

In the \textbf{E-step}, the expected complete-data log-likelihood function conditioned on the parameter $\boldsymbol{\theta}^{(t)}$ at the $t$-th iteration is given by:
\begin{align}
    \label{eq7}
    &\mathbb{E}_{\boldsymbol{Y}|\boldsymbol{X}, \boldsymbol{\theta}^{(t)}} \left\lbrack \log P(\boldsymbol{\theta}|\boldsymbol{X}, \boldsymbol{Y})\right\rbrack \\
    &=Const+\log\frac{1}{\boldsymbol{B}(\boldsymbol{\alpha})}+\sum_{j=1}^{K}(\alpha_j-1)\log\pi_j \nonumber \\
    &+\sum_{j=1}^{K}\sum_{i=1}^{N}P(Y_t=y_i=j|X_t=x_i,\boldsymbol{\theta}^{(t)})\log\pi_j \nonumber
\end{align}
where $Const$ includes all terms that are irrelevant to $\boldsymbol{\theta}$, such as $1/Z$ and $\sum_{i=1}^{N}P(X_s=x_i|Y_s=y_i)$. The detailed proof of (\ref{eq7}) is given in Appendix \ref{Appendix-1}.

Based on Bayes’ theorem and the label shift assumption defined in (\ref{eq1}), the conditional probability $P(Y_t=y_i=j \mid X_t=x_i, \boldsymbol{\theta}^{(t)})$ can be reformulated as \cite{Saerens2002}:
\begin{align}
    \label{eq8}
    P(Y_t=&y_i=j| X_t=x_i, \boldsymbol{\theta}^{(t)})\\
    &=\frac{\frac{\pi_{j}^{(t)}}{\varepsilon_j}P(Y_s=y_i=j|X_s=x_i)}{\sum_{k=1}^{K}\frac{\pi_{k}^{(t)}}{\varepsilon_k}P(Y_s=y_i=k|X_s=x_i)} = B_i^{(t)}\pi_{j}^{(t)} \nonumber 
\end{align}
where $\boldsymbol{\varepsilon} = (\varepsilon_1, \ldots, \varepsilon_K)$ denotes the training dataset's class prior probability vector.

For the update of $\boldsymbol{\pi}$ and $\boldsymbol{\alpha}$ in the \textbf{M-step}, the closed-formed $\boldsymbol{\pi}$ update function and the gradient ascent (GA)-based $\boldsymbol{\alpha}$ update function can be respectively written as:
\begin{align}
    \label{eq9}
    \pi_{j}^{(t+1)}=\frac{\alpha_j^{(t)}-1+\sum_{i=1}^{N}B_i^{(t)}\pi_j^{(t)}}{\sum_{k=1}^{K}\left(\alpha_k^{(t)}-1+\sum_{i=1}^{N}B_i^{(t)}\pi_k^{(t)}\right)}
\end{align}
\vspace{-8pt}
\begin{align}
    \label{eq10}
    \alpha_j^{(t+1)}=\alpha_j^{(t)}+\mu\left[-\psi(\alpha_j^{(t)})+\psi(\sum_{n=1}^{K}\alpha_n^{(t)})+\log\pi_j^{(t+1)}\right]
\end{align}
where $\psi(\cdot)$ is digamma function, $\mu$ is the learning rate. The detailed proof and the concavity analysis are shown in our prior work \cite{Hu2025}.

Next, we consider the CRB for prior probability $\bm{\pi}$ estimation. Assume $\hat{\bm{\pi}}$ is the estimated value of true prior probability, the estimation covariance can be written as:
\begin{align}
    \label{eq11}
    \text{Cov}\left(\hat{\bm{\pi}}\right) \geq C^{\text{off}}_{\text{CRB}}(\bm{\pi})
\end{align}
where $C^{\text{off}}_{\text{CRB}}(\bm{\pi})$ denotes the CRB for $\bm{\pi}$ estimation using the batch EM algorithm. According to statistical signal processing theory \cite{Kay1993}, the CRB provides the theoretical lower bound on the covariance of unbiased estimator, which can be expressed as the inverse of the Fisher information matrix (FIM) $\bm{F}$ corresponding to the estimated parameters \cite{Zhao2024}:
\begin{align}
    \label{eq12}
    C^{\text{off}}_{\text{CRB}}(\bm{\pi})=\bm{F}^{-1}
\end{align}
\begin{align}
    \label{eq13}
    F_{i,j}^{\text{off}}=\frac{\alpha_i-1+\sum_{n=1}^Nf_i^{x_n}}{\pi_i^2}\delta_{i,j} +\frac{\alpha_K-1+\sum_{n=1}^Nf_K^{x_n}}{\pi_K^2}
\end{align}
where $F_{i,j}^{\text{off}}$ are the elements of FIM, $f_{i}^{x_n}$ represents the posterior probability that the trained classifier classifies the input data $x_n$ as class $i$, $\delta_{i,j}$ is the Kronecker delta function. $\delta_{i,j}=1$ when $i=j$; otherwise, $\delta_{i,j}=0$. The detailed proof of FIM elements is shown in Appendix \ref{Appendix-2}. Therefore, the CRB of FMAPLS algorithm can be expressed as:
\begin{align}
    \label{eq14}
    C^{\text{off}}_{\text{CRB}}(\bm{\pi})=&\left(\text{diag}\Bigg[\frac{\alpha_i-1+\sum_{n=1}^Nf_i^{x_n}}{\pi_i^2} \right] \\
    &+\frac{\alpha_K-1+\sum_{n=1}^Nf_K^{x_n}}{\pi_K^2}\bm{1} \bm{1}^T\Bigg)^{-1} \nonumber
\end{align}
where $\text{diag}[\cdot]$ and $\bm{1} \bm{1}^T$ respectively represents a $(K-1)$-dimensional diagonal matrix and all-1 matrix.

\vspace{-12pt}
\subsection{Linear surrogate function (LSF)} \label{Section3-2}
Although GA can be employed to optimize $\boldsymbol{\alpha}$, its convergence speed and numerical stability are highly sensitive to both the choice of initialization and the learning rate $\mu$ \cite{McLachlan2007}. To mitigate these issues, this work introduces a LSF to replace equation (\ref{eq10}), thereby eliminating the sensitivity to the initialization and the learning rate of $\bm{\alpha}$ update step:
\begin{align}
    \label{eq15}
    \alpha_{j} = \frac{c}{\max(\bm{\pi}^{(t+1)})}\pi_{j}^{(t+1)}=\hat c\pi_{j}^{(t+1)}
\end{align}
where LSF parameter $c$ is a large constant. To demonstrate the rationality of the LSF method, it is necessary to prove that both GA and the surrogate scheme drive the partial derivative of the $\alpha$-related term in equation (\ref{eq7}) $L_{\bm{\alpha}}$ to converge to zero:
\begin{align}
    \label{eq16}
    L_{\bm{\alpha}}=\log\frac{1}{\boldsymbol{B}(\boldsymbol{\alpha})}+\sum_{j=1}^{K}(\alpha_j-1)\log\pi_j^{(t+1)}
\end{align}
Owing to the concavity of $\bm{\alpha}$ update function, employing GA with an appropriately selected $\mu$ ensures convergence to the maximum point \cite{Hu2025}. In the case of the LSF method, substituting (\ref{eq15}) into the partial derivative equation of $L_{\bm{\alpha}}$ leads to the following formula:
\begin{align}
    \label{eq17}
    \frac{\partial L_{\bm{\alpha}}}{\partial \alpha_j}&= -\psi(\hat c\pi_{j}^{ (t+1)})+\psi(\hat c)+\log\pi_j^{(t+1)} \\
    &=\left[\log(\hat c\pi_{j}^{(t+1)})-\psi(\hat c\pi_{j}^{(t+1)})\right]+\left[\psi(\hat c)-\log(\hat c)\right] \nonumber 
\end{align}
Based on the asymptotic expansion of the digamma function \cite{Bernardo1976}, equation (\ref{eq17}) can be further expressed as:
\begin{align}
    \label{eq18}
    &\left[\log(\hat c\pi_{j}^{(t+1)})-\psi(\hat c\pi_{j}^{(t+1)})\right]+\left[\psi(\hat c)-\log(\hat c)\right] \\
    &\sim \mathcal{O}(\frac{1}{\hat c})-\mathcal{O}(\frac{1}{\hat c\pi_{j}^{(t+1)}}) \nonumber \\
    &\sim \mathcal{O}(\frac{1}{\hat c\pi_{j}^{(t+1)}}) \nonumber
\end{align}
It can be shown that equation (\ref{eq18}) converges to zero as $c \to +\infty$. However, for practical implementations where $c$ is finite, the approximation error introduced by the surrogate function can be rigorously bounded within $\mathcal{O}(\frac{1}{\hat c\pi_{j}^{(t+1)}})$. Consequently, the surrogate update rule is effectively equivalent to performing GA for maximizing $L_{\bm{\alpha}}$ when $c$ is sufficiently large.

As a result, the batch EM procedure produces a parameter sequence $\{(\boldsymbol{\pi}^{(t)}, \boldsymbol{\alpha}^{(t)})\}$ with monotonically non-decreasing log-likelihood values, and the iteration continues until the maximum loop number $T_{\text{max}}$ is reached. When FMAPLS finishes estimating, the final estimation of the test prior distribution, $\boldsymbol{\pi}^*$, is then employed to reweight the classifier’s output soft posterior probabilities according to (\ref{eq8}). The overall procedure of FMAPLS is summarized in Algorithm~\ref{alg1}.

\begin{algorithm}[t]
    \caption{FMAPLS}
    \label{alg1}
    \renewcommand{\algorithmicrequire}{\textbf{Input:}}
    \renewcommand{\algorithmicensure}{\textbf{Output:}}
    
    \begin{algorithmic}[1]
        \REQUIRE Test dataset $\bm{X}$, training prior $\bm{\varepsilon}$ and classifier $f$
        \ENSURE Test domain prior probability $P(\bm{Y}_t)=\boldsymbol{\pi}^*$

        \STATE \textbf{Initialization Step}: Initialize Dirichlet parameter $\bm{\alpha}^{(0)}$ and test prior probability $\bm{\pi}^{(0)}$ as uniform distribution:
        \begin{align}
            \bm{\alpha}^{(0)}=\bm{1} \;\;\;\;\; \pi_j^{(0)}=\frac{1}{K} \nonumber
            \end{align}
        \FOR{$t=0$ to $T_{\text{max}}$}
            \STATE \textbf{E Step}: Calculate $\mathbb{E}_{\boldsymbol{Y}|\boldsymbol{X}, \boldsymbol{\theta}^{(t)}} \left\lbrack \log P(\boldsymbol{\theta}|\boldsymbol{X}, \boldsymbol{Y})\right\rbrack$ according to equation (\ref{eq7})
            \STATE \textbf{M Step}: Obtain $\bm{\pi}^{(t+1)}$ and $\bm{\alpha}^{(t+1)}$ via:
            \begin{align}
            \pi_{j}^{(t+1)}=\frac{\alpha_j^{(t)}-1+\sum_{i=1}^{N}B_i^{(t)}\pi_j^{(t)}}{\sum_{k=1}^{K}\left(\alpha_k^{(t)}-1+\sum_{i=1}^{N}B_i^{(t)}\pi_k^{(t)}\right)} \nonumber
            \end{align}
            \begin{align}
            \alpha_{j}^{(t+1)} = \frac{c}{\max(\bm{\pi}^{(t+1)})}\pi_{j}^{(t+1)}=\hat c\pi_{j}^{(t+1)} \nonumber
            \end{align}
        \ENDFOR
        \RETURN Test domain prior probability estimation: \\ $P(\bm{Y}_t)=\bm{\pi}^{T_{\text{max}}}=\boldsymbol{\pi}^*$
        \STATE Reweight classifier output according to (\ref{eq8})
    \end{algorithmic}
\end{algorithm}

\vspace{-8pt}
\section{Online EM Method} \label{Section4}
This section presents the online EM-based test prior estimation algorithm, referred to as online-FMAPLS. Particularly, at time $\tau$, we can only observe classifier’s soft probability output $\bm{P}(\bm{Y}|X^{\tau})$, and the algorithm is able to process it in an online manner.

\subsection{Problem setting and assumption validation} \label{Section4-1}
Assume the neural network-based classifier have an input data $\boldsymbol{X}=x^{\tau}$ at time $t=\tau$. Similar to (\ref{eq6}), the complete posterior probability of parameter set $\bm{\theta}$ can be written as:
\begin{align}
    \label{eq19}
    P(\boldsymbol{\theta}|\boldsymbol{X}, \boldsymbol{Y})&=\frac{P(\boldsymbol{X}, \boldsymbol{Y}|\boldsymbol{\theta})P(\boldsymbol{\theta})}{\int_{\boldsymbol{\theta}}P(\boldsymbol{X}, \boldsymbol{Y}|\boldsymbol{\theta})d\boldsymbol{\theta}} \\
    &=\frac{1}{Z}P(\boldsymbol{\theta})P(X_t=x^{\tau}, Y_t=y^{\tau}|\boldsymbol{\theta}) \nonumber \\
    &=\frac{1}{Z}P(\boldsymbol{\theta})P(X_t=x^{\tau}|Y_t=y^{\tau})P(Y_t=y^{\tau}|\boldsymbol{\theta}) \nonumber \\
    &=\frac{1}{Z}P(\boldsymbol{\theta})P(X_s=x^{\tau}|Y_s=y^{\tau})P(Y_t=y^{\tau}|\boldsymbol{\theta}) \nonumber
\end{align}

Different from the batch EM, its online variant replaces the E-step by a stochastic approximation step, while maintaining the M-step unchanged \cite{Wang2023}. According to the discussion in \cite{Cappé2009}, the following three assumptions need to be satisfied:

(1) \textit{Exponential form complete posterior}: We first prove that the complete posterior probability $P(\boldsymbol{\theta}|\boldsymbol{X}, \boldsymbol{Y})$ belongs to the \textit{exponential family} \cite{Nielsen2010}:
\begin{align}
    \label{eq20}
    P(\boldsymbol{\theta}|\boldsymbol{X}, \boldsymbol{Y})=h(\boldsymbol{X}, \boldsymbol{Y})\exp\{-\psi(\boldsymbol{\theta})+\langle \bm{S}(\boldsymbol{X}, \boldsymbol{Y}), \bm{\phi}(\boldsymbol{\theta})\rangle\}
\end{align}
\vspace{-14pt}
\begin{align}
    \label{eq21}
    \left\{
        \begin{aligned}
        &h(\boldsymbol{X},\boldsymbol{Y})=\frac{1}{Z}P(X_s=x^\tau|Y_s=y^\tau)\\
        &\psi(\boldsymbol{\theta})=\log(\boldsymbol{B}(\boldsymbol{\alpha}))\\
        &S_i(\boldsymbol{X},\boldsymbol{Y})=\begin{pmatrix}
                  1 \\
                  \delta_{y^\tau,i}
              \end{pmatrix} \\
        &\phi_i(\boldsymbol{\theta})=\begin{pmatrix}
                  (\alpha_i-1)\log\pi_i \\
                  \log\pi_i
              \end{pmatrix}
    \end{aligned}
    \right.
\end{align}
where $\langle\cdot, \cdot\rangle$ is the scalar product between two vectors, $\delta_{y^\tau,i}$ is the indicator function: $\delta_{y^\tau,i}=1$ if $y^\tau=i$ and $\delta_{y^\tau,i}=0$ otherwise. The detailed proof of (\ref{eq21}) is given in Appendix \ref{Appendix-3}.

(2) \textit{Well-defined $\bm{S}(\bm{X},\bm{Y})$ conditional expectation}: The conditional expectation of \textit{sufficient statistic} $\bm{S}(\bm{X},\bm{Y})$ with input data $x^{\tau}$ can be expressed as:
\begin{align}
    \label{eq22}
    \bar{S}_i(x^{\tau}; \boldsymbol{\theta})&=\mathbb{E}_{\bm{\theta}}[S_{i}(\bm{X},\bm{Y})|\bm{X}=x^{\tau}] \\
    &=\begin{pmatrix}
            1 \\
            E_{\bm{\theta}}[\delta_{y^\tau,i}|X_t=x^{\tau}]
        \end{pmatrix} \nonumber \\
    &=\begin{pmatrix}
            1 \\
            P(Y_t=y^{\tau}=i| X_t=x^{\tau}, \boldsymbol{\theta})
        \end{pmatrix} \nonumber \\
    &=\begin{pmatrix}
            1 \\
            B_i^{\tau}\pi_{i}
        \end{pmatrix} \nonumber
\end{align}
where $B_i^{\tau}\pi_{i}$ is defined in (\ref{eq8}). Since the classifier structure remains fixed during testing phase and the parameter $\boldsymbol{\theta}$ is treated as a known quantity in the expectation process, equation (\ref{eq22}) is consequently well-defined.

(3) \textit{Uniqueness of the maximum}: Ensuring an unambiguous M-step in the iterative process requires proving the following function possesses a unique maximizer:
\begin{align}
    \label{eq23}
    l(\bm{S},\bm{\theta})&=-\psi(\boldsymbol{\theta})+\langle \bm{S}(\boldsymbol{X}, \boldsymbol{Y}), \bm{\phi}(\boldsymbol{\theta})\rangle \\
    &=\log\left(\frac{1}{\boldsymbol{B}(\boldsymbol{\alpha})}\right)+\sum_{i=1}^K(\alpha_i-1+\delta_{y^\tau,i})\log\pi_i \nonumber
\end{align}
Since online-FMAPLS updates $\bm{\pi}$ and $\bm{\alpha}$ sequentially, it is further necessary to prove that $l(\bm{S},\bm{\theta})$ is concave in $\boldsymbol{\pi}$ when $\boldsymbol{\alpha}$ is fixed, and vice versa. The concavity with respect to $\bm{\pi}$ follows from the fact that $\log \pi_i$ is strictly concave on the probability simplex with nonnegative coefficients $(\alpha_i-1+\delta_{y^\tau,i})$, while for fixed $\bm{\pi}$, the term $-\log \boldsymbol{B}(\bm{\alpha})$ is strictly concave in $\bm{\alpha}$ according to the discussion \cite{Hu2025, Nielsen2010, Huang2005}. Therefore, $l(\bm{S},\bm{\theta})$ is separately concave in $\bm{\pi}$ and $\bm{\alpha}$, ensuring the uniqueness of the maximum in each update step.

\vspace{-12pt}
\subsection{Online-FMAPLS algorithm} \label{Section4-2}
According to the procedure of the online EM \cite{Cappé2009}, the \textbf{online E-step} takes the following form:
\begin{align}
    \label{eq24}
    S_{i}^{\tau+1}&=S_{i}^{\tau}+\gamma\left\{\bar{S}_i(x^{\tau+1}; \boldsymbol{\theta}^{\tau})-S_{i}^{\tau}\right\} \\
    &=\begin{pmatrix}
        1 \\
        (1-\gamma)\delta_{y^\tau,i}+\gamma B_i^{\tau+1}\pi_{i}^{\tau}
    \end{pmatrix} \nonumber
\end{align}
where $\gamma$ denotes the confidence parameter of the online label shift estimation framework. As $\gamma \to 1^{-}$, the algorithm places greater emphasis on future data, whereas as $\gamma \to 0^{+}$, it focuses more heavily on the current data.

In the procedure of the \textbf{M step} in online EM algorithm, the goal is to maximize the following equation:
\begin{align}
    \label{eq25}
    &\{\boldsymbol{\pi}^{\tau+1}, \boldsymbol{\alpha}^{\tau+1}\}=\mathop{\arg\max}_{\boldsymbol{\theta}^{\tau+1}}\;l(\bm{S}^{\tau+1}, \bm{\theta}) \\
    &=\mathop{\arg\max}_{\boldsymbol{\theta}^{\tau+1}}\;\log\frac{1}{\boldsymbol{B}(\boldsymbol{\alpha})} \nonumber \\
    &+\left\langle \begin{pmatrix}
            1 \\
            (1-\gamma)\delta_{y^\tau}+\gamma B^{\tau+1}\pi^{\tau}
        \end{pmatrix}_i, \begin{pmatrix}
            (\alpha-1)\log\pi \nonumber \\
            \log\pi
        \end{pmatrix}_i \right\rangle\\
    &=\mathop{\arg\max}_{\boldsymbol{\theta}^{\tau+1}}\;\log\frac{1}{\boldsymbol{B}(\boldsymbol{\alpha})}+\sum_{i=1}^K\big\{(\alpha_{i}-1)\log\pi_{i} \nonumber \\
    &+\big[(1-\gamma)\delta_{y^\tau,i}+\gamma B_i^{\tau+1}\pi_{i}^{\tau}\big]\log\pi_{i}\big\} \nonumber
\end{align}

For the update of $\boldsymbol{\pi}$, the goal is to maximize terms related to $\boldsymbol{\pi}$ in the maximize function with normalization constraint, which can be expressed as:
\begin{align}
    \label{eq26}
    \begin{cases}
        \displaystyle 
        \mathop{\arg\max}_{\boldsymbol{\pi}^{\tau+1}} \sum_{i=1}^K\big\{(\alpha_{i}^{\tau}-1)\log\pi_{i}\\
        +\big[(1-\gamma)\delta_{y^\tau,i}+\gamma B_i^{\tau+1}\pi_{i}^{\tau}\big]\log\pi_{i}\big\} \\[2mm]
        \text{s.t.}: \;\;\;\;\; \sum_{j=1}^{K} \pi_j = 1,\ \pi_j > 0
    \end{cases}
\end{align}
Using the Lagrange Multiplier Method, the Lagrangian equation $ L_{\pi}^{\text{on}}(\boldsymbol{\pi},\lambda)$ can be written as:
\begin{align}
    \label{eq27}
    L_{\pi}^{\text{on}}(\boldsymbol{\pi},\lambda)&=\sum_{i=1}^K\big\{(\alpha_{i}^{\tau}-1)\log\pi_{i}+\big[(1-\gamma)\delta_{y^\tau,i} \\
    &+\gamma B_i^{\tau+1}\pi_{i}^{\tau}\big]\log\pi_{i}\big\}+\lambda\left(1-\sum_{k=1}^{K}\pi_k\right) \nonumber
\end{align}
where $\lambda$ is a Lagrange multiplier. Computing the partial derivatives of (\ref{eq27}) with respect to $\pi_i$ and $\lambda$ leads to the following set of equations:
\begin{align}
    \label{eq28}
    \frac{\partial L_{\pi}^{\text{on}}}{\partial \pi_i}=\frac{(\alpha_i^{\tau}-1)+(1-\gamma)\delta_{y^\tau,i}+\gamma B_i^{\tau+1}\pi_i^{\tau}}{\pi_i}-\lambda=0
\end{align}
\vspace{-12pt}
\begin{align}
    \label{eq29}
    \frac{\partial L_{\pi}^{\text{on}}}{\partial \lambda}=1-\sum_{k=1}^{K}\pi_k=0
\end{align}
Substituting (\ref{eq28}) into (\ref{eq29}), $\lambda$ can be given as:
\begin{align}
    \label{eq30}
    \lambda=\sum\limits_{k=1}^{K}\left((\alpha_k^{\tau}-1)+(1-\gamma)\delta_{y^\tau,k}+\gamma B_k^{\tau+1}\pi_k^{\tau}\right)
\end{align}
Substituting (\ref{eq30}) into (\ref{eq28}), the update formula for $\pi_{i}^{\tau+1}$ can finally be expressed as:
\begin{align}
    \label{eq31}
    \pi_{i}^{\tau+1}=\frac{(\alpha_i^{\tau}-1)+(1-\gamma)\delta_{y^\tau,i}+\gamma B_i^{\tau+1}\pi_i^{\tau}}{\sum\limits_{k=1}^{K}\left((\alpha_k^{\tau}-1)+(1-\gamma)\delta_{y^\tau,k}+\gamma B_k^{\tau+1}\pi_k^{\tau}\right)}
\end{align}

As for the update of $\bm{\alpha}$, the goal is to maximize the following equation:
\begin{align}
    \label{eq32}
    \mathop{\arg\max}_{\boldsymbol{\alpha}^{\tau+1}} \;\log\frac{1}{\boldsymbol{B}(\boldsymbol{\alpha})}+\sum_{i=1}^{K}(\alpha_i-1)\log\pi_i^{\tau+1}
\end{align}
Notably, (\ref{eq32}) has the same construct with the M-step of $\bm{\alpha}$ in batch EM. Thus, the update procedure of $\bm{\alpha}$ can also be replaced by LSF method:
\begin{align}
    \label{eq33}
    \alpha_{i}^{\tau+1} = \frac{c}{\max(\bm{\pi}^{\tau+1})}\pi_{i}^{\tau+1}=\hat c\pi_{i}^{\tau+1}
\end{align}
where $c$ is a sufficiently large constant. In summary, the pseudocode of online-FMAPLS is shown in Algorithm~\ref{alg2}.

Similar to the CRB analysis process in Section~\ref{Section3-1}, the FIM elements $F_{i,j}^{\text{on}}$ of online-FMAPLS can be given as:
\begin{align}
    \label{eq34}
    F_{i,j}^{\text{on}}&=\frac{\alpha_i^{\tau}-1+(1-\gamma)\delta_{y^{\tau},i}+\gamma\frac{1}{\tau+1}\sum\limits_{t=1}^{\tau+1} f_i^{x^t}}{\pi_i^2}\delta_{i,j} \\
    &+\frac{\alpha_K^{\tau}-1+(1-\gamma)\delta_{y^{\tau},K}+\gamma\frac{1}{\tau+1}\sum\limits_{t=1}^{\tau+1} f_K^{x^t}}{\pi_K^2} \nonumber
\end{align}
where the detailed proof is shown in Appendix~\ref{Appendix-4}. Thus, the CRB of online-FMAPLS is:
\begin{align}
    \label{eq35}
    &C_{\text{CRB}}^{on}(\bm{\pi})=\Bigg(\text{diag}\left[\frac{\alpha_i^{\tau}-1+(1-\gamma)\delta_{y^{\tau},i}+\gamma\frac{1}{\tau+1}\sum\limits_{t=1}^{\tau+1} f_i^{x^t}}{\pi_i^2} \right] \\
    &+\frac{\alpha_K^{\tau}-1+(1-\gamma)\delta_{y^{\tau},K}+\gamma\frac{1}{\tau+1}\sum\limits_{t=1}^{\tau+1} f_K^{x^t}}{\pi_K^2}\bm{1} \bm{1}^T\Bigg)^{-1} \nonumber
\end{align}
where $\text{diag}[\cdot]$ and $\bm{1} \bm{1}^T$ respectively represents a $(K-1)$-dimensional diagonal matrix and all-1 matrix.

\begin{algorithm}[t] 
    \caption{online-FMAPLS}
    \label{alg2}
    \renewcommand{\algorithmicrequire}{\textbf{Input:}}
    \renewcommand{\algorithmicensure}{\textbf{Output:}}
    
    \begin{algorithmic}[1]
        \REQUIRE Test dataset $\bm{X}$, training prior $\bm{\varepsilon}$ and classifier $f$
        \ENSURE Test domain prior probability $P(Y_t=\cdot)=\bm{\pi}^{N}$

        \STATE \textbf{Initialization Step}: Initialize Dirichlet parameter $\bm{\alpha}^{(0)}$ and test prior probability $\bm{\pi}^{(0)}$ as uniform distribution:
        \begin{align}
            \bm{\alpha}^{(0)}=\bm{1} \;\;\;\;\; \pi_j^{(0)}=\frac{1}{K} \nonumber
            \end{align}
        \FOR{$\tau=0$ to $N$}
            \STATE \textbf{Online E Step}: Calculate $\bm{S}^{\tau+1}$ function according to equation (\ref{eq24})
            \STATE \textbf{Online M Step}: Obtain $\bm{\alpha}^{\tau+1}$ and $\bm{\pi}^{\tau+1}$ via:
            \begin{align}
            \pi_{i}^{\tau+1}=\frac{(\alpha_i^{\tau}-1)+(1-\gamma)\delta_{y^\tau,i}+\gamma B_i^{\tau+1}\pi_i^{\tau}}{\sum\limits_{k=1}^{K}\left((\alpha_k^{\tau}-1)+(1-\gamma)\delta_{y^\tau,k}+\gamma B_k^{\tau+1}\pi_k^{\tau}\right)} \nonumber
            \end{align}
            \begin{align}
            \alpha_{i}^{\tau+1} = \frac{c}{\max(\bm{\pi}^{\tau+1})}\pi_{i}^{\tau+1}=\hat c\pi_{i}^{\tau+1} \nonumber
            \end{align}
        \ENDFOR
        \RETURN Test domain prior probability estimation: \\ $P(\bm{Y}_t)=\bm{\pi}^{N}$
        \STATE Reweight classifier output according to (\ref{eq8})
    \end{algorithmic}
\end{algorithm}

\vspace{-10pt}
\subsection{Convergence rate analysis} \label{Section4-3}
In streaming environments, where data arrive sequentially and computational resources are limited, rapid convergence of estimation algorithms is critically important. The ability to quickly adapt to label shift with few samples directly affects the practical utility of online systems. This motivates a detailed analysis of online-FMAPLS convergence, with emphasis on parameter $c$, which governs the trade-off between convergence rate and estimation accuracy.

Substituting the LSF into online $\bm{\pi}$ update function yields the following recursive form of the test prior $\pi_{i}^{\tau+1}$:
\begin{align}
    \label{eq36}
    \pi_{i}^{\tau+1}&=\frac{(\hat c\pi_{i}^{\tau}-1)+(1-\gamma)\delta_{y^\tau,i}+\gamma B_i^{\tau+1}\pi_i^{\tau}}{\sum\limits_{k=1}^{K}\left((\hat c\pi_{k}^{\tau}-1)+(1-\gamma)\delta_{y^\tau,k}+\gamma B_k^{\tau+1}\pi_k^{\tau}\right)} \\
    &=\frac{\hat c\pi_{i}^{\tau}+\gamma B_i^{\tau+1}\pi_i^\tau+(1-\gamma)\delta_{y^\tau,i}-1}{\hat c-K+1-\gamma+\gamma \sum\limits_{k=1}^{K}B_k^{\tau+1}\pi_k^{\tau}} \nonumber
\end{align}
Let $M=\sum_{k=1}^{K}B_k^{\tau+1}\pi_k^{\tau}$, the increment $|\pi_{i}^{\tau+1}-\pi_{i}^{\tau}|$ between two consecutive iterations can be consequently expressed as:
\begin{align}
    \label{eq37}
    &|\pi_{i}^{\tau+1}-\pi_{i}^{\tau}| \\
    &=\left|\frac{\pi_{i}^{\tau}(\gamma B_i^{\tau+1}-\gamma M+\gamma+K-1)+(1-\gamma)\delta_{y^\tau,i}-1}{\hat c-K+1-\gamma+\gamma M}\right| \nonumber
\end{align}
It is evident from (\ref{eq37}) that the magnitude of the iterative increment is inversely proportional to the term $\hat c$. Therefore, when parameters $K, M, \gamma$ are fixed during each iteration, the update step of $\pi_i$ satisfies:
\begin{align}
    \label{eq38}
    |\pi_{i}^{\tau+1}-\pi_{i}^{\tau}|\sim\mathcal{O}(\frac{1}{\hat c})
\end{align}

This analytical result shows that larger $c$ yields smaller update steps and consequently slower convergence, while smaller $c$ produces faster convergence. However, as discussed in Section~\ref{Section3-2}, the LSF introduces considerable estimation errors when $c$ is relatively small, causing a gradual drift in the estimated test priors. Therefore, online-FMAPLS exhibits a fundamental trade-off between convergence rate and estimation accuracy, which is governed by $c$. In streaming scenarios with sequential data and limited resources, rapid convergence is often prioritized to quickly adapt to distribution shifts. This necessitates accepting a certain level of estimation error in exchange for improved responsiveness.

\vspace{-10pt}
\section{Complexity Analysis} \label{Section5}
\vspace{-3pt}
\begin{table}[b]
\vspace{-22pt}
\caption{Computational Complexity of Each Iteration of the Algorithms\label{Table1}}
\vspace{-5pt}
\centering
{\scriptsize
\setlength{\tabcolsep}{15pt}
\begin{tabular}{|c|c|}
\hline
Algorithm & Computational Complexity\\
\hline
FMAPLS+GA & $\mathcal{O}(NK+T_{\text{grad}}K)$\\
\hline
FMAPLS+LSF & $\mathcal{O}(NK+K)$\\
\hline
online-FMAPLS+GA & $\mathcal{O}(K+T_{\text{grad}}K)$\\
\hline
online-FMAPLS+LSF & $\mathcal{O}(K)$\\
\hline
\end{tabular}
}
\end{table}
This section investigates the computational complexity of proposed methods in a single EM iteration. Each iteration requires the calculation of expectation formula followed by sequential updates of $K$-dimensional vectors $\bm{\pi}$ and $\bm{\alpha}$: 

(1) Offline expectation: As shown in equation (\ref{eq7}), the computational cost of expectation formula requires $\mathcal{O}(NK)$ in total. 

(2) Offline $\bm{\pi}$ update: Equation (\ref{eq9}) shows that, the complexity of update $\pi_j^{(t+1)}$ requires $\mathcal{O}(NK)$.

(3) Offline $\bm{\alpha}$ update with GA method: Equation (\ref{eq10}) shows that, the complexity of update $\alpha_j^{(t+1)}$ requires $\mathcal{O}(T_{\text{grad}}K)$, where $T_{\text{grad}}$ is the iteration number of each GA.

(4) Offline $\bm{\alpha}$ update with LSF method: Due to the linear property, the computing complexity of equation (\ref{eq15}) is $\mathcal{O}(K)$.

(5) Online expectation: Since data is input in a streaming manner, the computational complexity of the online expectation formula as shown in equation (\ref{eq24}) is $\mathcal{O}(K)$.

(6) Online $\bm{\pi}$ update: Equation (\ref{eq31}) shows that, the complexity of update $\pi_i^{\tau+1}$ requires $\mathcal{O}(K)$.

(7) Online $\bm{\alpha}$ update: The same as the offline $\bm{\alpha}$ update analysis, the computational complexity of the gradient ascent and LSF solutions are $\mathcal{O}(T_{\text{grad}}K)$ and $\mathcal{O}(K)$ respectively.

In summary, the computational complexities of the four proposed methods are listed in Table~\ref{Table1}. For updating $\boldsymbol{\alpha}$, GA requires $T_{\text{grad}}$ iterations, leading to a cost of $\mathcal{O}(T_{\text{grad}}K)$, whereas LSF performs a single scaling operation with complexity $\mathcal{O}(K)$. Comparing batch-based FMAPLS with its streaming version, online-FMAPLS reduces the cost of expectation and $\boldsymbol{\pi}$ update steps from $\mathcal{O}(NK)$ to $\mathcal{O}(K)$ by replacing batch aggregation with streaming updates. Overall, LSF is more efficient than GA for updating $\boldsymbol{\alpha}$, and online-FMAPLS is better suited for large-scale or streaming applications.

\vspace{-12pt}
\section{Numerical Experiments} \label{Section6}
\vspace{-3pt}
\begin{table*}[!t]
\centering
\vspace{-10pt}
\caption{KL Divergence and Classification Accuracy (“KL$|$Acc.”) on Shuffled Long-Tail Test Distributions with Training Imbalance Ratios $\{0.2, 0.1, 0.05, 0.02\}$ and Test Ratios $\{0.02, 0.025, 0.04, 0.05, 0.1\}$. Best Performances are in Bold and Second Best in Blue.}
\label{Table2}
\vspace{-8pt}
{\scriptsize
\setlength{\tabcolsep}{4.8pt}
\begin{tabular}{c|ccccc|ccccc} % 指定表格列数和对齐方式
\toprule % 顶部横线
Dataset & \multicolumn{5}{c}{CIFAR100-0.2-LT} & \multicolumn{5}{|c}{CIFAR100-0.1-LT}\\ % 表头
\midrule % 中间横线
Imbalance ratio $\rho_{\text{test}}$ & 0.02 & 0.025 & 0.04 & 0.05 & 0.1 & 0.02 & 0.025 & 0.04 & 0.05 & 0.1 \\
\midrule % 中间横线
MLLS & 0.109$|$\textbf{68.48} & 0.101$|$\textbf{68.32} & 0.075$|$\textbf{67.24} & 0.066$|$\textbf{67.14} & 0.037$|$65.98 & 0.190$|$\textbf{63.42} & 0.178$|$\textcolor{blue}{63.02} & 0.135$|$\textcolor{blue}{62.30} & 0.120$|$\textcolor{blue}{62.02} & 0.083$|$60.78 \\
BBSE & \textcolor{blue}{0.069}$|$68.31 & \textcolor{blue}{0.062}$|$68.15 & 0.057$|$67.10 & \textcolor{blue}{0.050}$|$67.05 & \textcolor{blue}{0.036}$|$\textcolor{blue}{66.00} & 0.145$|$63.25 & 0.131$|$63.00 & 0.102$|$62.24 & 0.098$|$62.01 & 0.067$|$60.91 \\
RLLS & 0.606$|$62.98 & 0.558$|$63.35 & 0.488$|$62.92 & 0.450$|$63.25 & 0.344$|$63.10 & 0.740$|$57.62 & 0.697$|$57.87 & 0.623$|$57.58 & 0.580$|$57.87 & 0.484$|$57.59 \\
MAPLS & \textcolor{blue}{0.069}$|$68.22 & 0.063$|$68.06 & \textcolor{blue}{0.056}$|$67.08 & 0.052$|$67.03 & 0.041$|$65.98 & \textcolor{blue}{0.103}$|$63.17 & \textcolor{blue}{0.097}$|$62.94 & \textcolor{blue}{0.086}$|$62.17 & \textcolor{blue}{0.080}$|$61.94 & \textcolor{blue}{0.064}$|$\textcolor{blue}{60.94} \\
\midrule % 中间横线
\textbf{FMAPLS} & \textbf{0.053}$|$\textcolor{blue}{68.32} & \textbf{0.048}$|$\textcolor{blue}{68.16} & \textbf{0.041}$|$\textcolor{blue}{67.18} & \textbf{0.038}$|$\textcolor{blue}{67.09} & \textbf{0.028}$|$\textbf{66.05} & \textbf{0.079}$|$\textcolor{blue}{63.32} & \textbf{0.073}$|$\textbf{63.03} & \textbf{0.063}$|$\textbf{62.34} & \textbf{0.058}$|$\textbf{62.07} & \textbf{0.046}$|$\textbf{60.96} \\
\textbf{online-FMAPLS} & 0.082$|$68.11 & 0.078$|$67.95 & 0.071$|$66.90 & 0.067$|$66.90 & 0.057$|$65.72 & 0.108$|$63.13 & 0.103$|$62.89 & 0.092$|$62.10 & 0.087$|$61.84 & 0.075$|$60.69\\
\midrule % 中间横线
\midrule % 中间横线
Dataset & \multicolumn{5}{c}{CIFAR100-0.05-LT} & \multicolumn{5}{|c}{CIFAR100-0.02-LT}\\ % 表头
\midrule % 中间横线
Imbalance ratio $\rho_{\text{test}}$ & 0.02 & 0.025 & 0.04 & 0.05 & 0.1 & 0.02 & 0.025 & 0.04 & 0.05 & 0.1 \\
\midrule % 中间横线
MLLS & 0.196$|$\textbf{60.75} & 0.181$|$\textbf{60.21} & 0.156$|$\textcolor{blue}{58.76} & 0.118$|$\textcolor{blue}{58.91} & 0.091$|$\textcolor{blue}{57.85} & 0.339$|$\textbf{52.93} & 0.308$|$\textbf{52.73} & 0.307$|$\textcolor{blue}{51.41} & 0.293$|$\textcolor{blue}{51.20} & 0.231$|$\textcolor{blue}{50.38} \\
BBSE & 0.268$|$60.19 & 0.250$|$59.68 & 0.238$|$58.38 & 0.209$|$58.48 & 0.117$|$57.68 & 2.264$|$45.06 & 0.646$|$51.99 & 0.914$|$49.82 & 1.998$|$44.94 & 1.165$|$48.03 \\
RLLS & 0.909$|$52.35 & 0.860$|$52.29 & 0.803$|$51.59 & 0.770$|$51.79 & 0.652$|$52.17 & 1.152$|$44.38 & 1.129$|$44.08 & 1.074$|$43.55 & 1.015$|$44.07 & 0.931$|$44.12 \\
MAPLS & 0.129$|$60.25 & 0.122$|$59.71 & 0.109$|$58.47 & 0.104$|$58.53 & \textcolor{blue}{0.079}$|$57.78 & 0.182$|$52.24 & 0.171$|$51.97 & 0.151$|$50.88 & 0.138$|$50.74 & 0.102$|$50.19 \\
\midrule % 中间横线
\textbf{FMAPLS} & \textbf{0.085}$|$\textcolor{blue}{60.62} & \textbf{0.079}$|$\textcolor{blue}{60.09} & \textbf{0.070}$|$\textbf{58.78} & \textbf{0.062}$|$\textbf{58.94} & \textbf{0.047}$|$\textbf{58.02} & \textbf{0.124}$|$\textcolor{blue}{52.81} & \textbf{0.113}$|$\textcolor{blue}{52.62} & \textbf{0.099}$|$\textbf{51.46} & \textbf{0.089}$|$\textbf{51.32} & \textbf{0.062}$|$\textbf{50.59} \\
\textbf{online-FMAPLS} & \textcolor{blue}{0.122}$|$60.26 & \textcolor{blue}{0.115}$|$59.78 & \textcolor{blue}{0.106}$|$58.44 & \textcolor{blue}{0.099}$|$58.50 & 0.085$|$57.52 & \textcolor{blue}{0.161}$|$52.66 & \textcolor{blue}{0.149}$|$52.32 & \textcolor{blue}{0.138}$|$51.19 & \textcolor{blue}{0.130}$|$50.95 & \textcolor{blue}{0.101}$|$\textcolor{blue}{50.38} \\
\bottomrule % 底部横线
\end{tabular}
}
\end{table*}

\begin{table*}[!t]
\vspace{-12pt}
\centering
\caption{KL Divergence and Classification Accuracy (“KL$|$Acc.”) on Dirichlet Test Distributions with Training Imbalance Ratios $\{0.2, 0.1, 0.05, 0.02\}$ and Test Dirichlet Hyperparameters $\{1, 1.5, 2, 2.5, 3\}$. Best Performances are in Bold and Second Best in Blue.}
\label{Table3}
\vspace{-8pt}
{\scriptsize
\setlength{\tabcolsep}{4.8pt}
\begin{tabular}{c|ccccc|ccccc} % 指定表格列数和对齐方式
\toprule % 顶部横线
Dataset & \multicolumn{5}{c|}{CIFAR100-0.2-LT} & \multicolumn{5}{c}{CIFAR100-0.1-LT} \\
\midrule % 中间横线
Dirichlet $\bm{\alpha}_{\text{test}}$ & 1 & 1.5 & 2 & 2.5 & 3 & 1 & 1.5 & 2 & 2.5 & 3 \\
\midrule % 中间横线
MLLS & 0.082$|$\textbf{68.05} & 0.067$|$\textbf{66.83} & 0.057$|$66.06 & 0.045$|$65.91 & 0.046$|$65.33 & 0.144$|$\textbf{62.84} & 0.122$|$\textcolor{blue}{61.72} & 0.120$|$60.62 & 0.098$|$60.32 & 0.102$|$60.20 \\
BBSE & \textcolor{blue}{0.062}$|$67.91 & \textcolor{blue}{0.052}$|$66.75 & \textcolor{blue}{0.046}$|$\textcolor{blue}{66.09} & \textcolor{blue}{0.038}$|$\textcolor{blue}{65.96} & \textcolor{blue}{0.038}$|$65.42 & 0.115$|$62.68 & 0.096$|$61.65 & 0.088$|$60.71 & 0.082$|$60.46 & 0.083$|$60.34 \\
RLLS & 0.541$|$63.23 & 0.423$|$63.16 & 0.368$|$63.07 & 0.320$|$63.52 & 0.300$|$63.12 & 0.689$|$57.53 & 0.564$|$57.67 & 0.500$|$57.45 & 0.461$|$57.58 & 0.431$|$57.74 \\
MAPLS & 0.065$|$67.77 & 0.054$|$66.69 & 0.047$|$66.08 & 0.041$|$\textcolor{blue}{65.96} & \textcolor{blue}{0.038}$|$\textcolor{blue}{65.43} & \textcolor{blue}{0.097}$|$62.57 & \textcolor{blue}{0.080}$|$61.66 & \textcolor{blue}{0.068}$|$\textcolor{blue}{60.81} & \textcolor{blue}{0.061}$|$\textcolor{blue}{60.52} & \textcolor{blue}{0.056}$|$\textcolor{blue}{60.45} \\
\midrule % 中间横线
\textbf{FMAPLS} & \textbf{0.043}$|$\textcolor{blue}{67.97} & \textbf{0.038}$|$\textcolor{blue}{66.81} & \textbf{0.034}$|$\textbf{66.12} & \textbf{0.030}$|$\textbf{65.99} & \textbf{0.029}$|$\textbf{65.45} & \textbf{0.066}$|$\textcolor{blue}{62.81} & \textbf{0.059}$|$\textbf{61.79} & \textbf{0.054}$|$\textbf{60.83} & \textbf{0.049}$|$\textbf{60.56} & \textbf{0.046}$|$\textbf{60.47} \\
\textbf{online-FMAPLS} & 0.076$|$67.61 & 0.061$|$66.61 & 0.054$|$65.97 & 0.044$|$65.92 & 0.042$|$65.36 & 0.100$|$62.53 & 0.084$|$61.61 & 0.073$|$60.70 & 0.064$|$60.50 & 0.059$|$\textcolor{blue}{60.45} \\
\midrule % 中间横线
\midrule % 中间横线
Dataset & \multicolumn{5}{c|}{CIFAR100-0.05-LT} & \multicolumn{5}{c}{CIFAR100-0.02-LT} \\
\midrule % 中间横线
Dirichlet $\bm{\alpha}_{\text{test}}$ & 1 & 1.5 & 2 & 2.5 & 3 & 1 & 1.5 & 2 & 2.5 & 3 \\
\midrule % 中间横线
MLLS & 0.146$|$\textbf{60.08} & 0.130$|$\textcolor{blue}{58.56} & 0.108$|$\textcolor{blue}{58.16} & 0.110$|$57.35 & 0.104$|$57.17 & 0.267$|$\textcolor{blue}{52.69} & 0.289$|$\textcolor{blue}{51.31} & 0.360$|$50.31 & 0.262$|$49.80 & 0.279$|$49.64 \\
BBSE & 0.188$|$59.74 & 0.164$|$58.38 & 0.132$|$57.79 & 0.150$|$57.12 & 0.183$|$56.69 & 1.277$|$50.10 & 0.761$|$50.69 & 3.136$|$40.57 & 0.922$|$48.56 & 0.642$|$49.33 \\
RLLS & 0.858$|$52.23 & 0.738$|$51.93 & 0.672$|$52.31 & 0.637$|$52.01 & 0.613$|$52.16 & 1.124$|$44.22 & 0.994$|$44.33 & 0.937$|$44.37 & 0.904$|$44.04 & 0.883$|$43.94 \\
MAPLS & 0.123$|$59.54 & 0.100$|$58.31 & 0.084$|$58.13 & 0.073$|$\textcolor{blue}{57.48} & 0.067$|$\textcolor{blue}{57.48} & 0.165$|$51.98 & 0.134$|$50.98 & \textcolor{blue}{0.109}$|$50.36 & 0.095$|$50.03 & \textcolor{blue}{0.084}$|$49.72 \\
\midrule % 中间横线
\textbf{FMAPLS} & \textbf{0.074}$|$\textcolor{blue}{59.99} & \textbf{0.063}$|$\textbf{58.63} & \textbf{0.053}$|$\textbf{58.31} & \textbf{0.048}$|$\textbf{57.67} & \textbf{0.046}$|$\textbf{57.50} & \textbf{0.102}$|$\textbf{52.71} & \textbf{0.088}$|$\textbf{51.41} & \textbf{0.077}$|$\textbf{50.74} & \textbf{0.065}$|$\textbf{50.25} & \textbf{0.059}$|$\textbf{50.00} \\
\textbf{online-FMAPLS} & \textcolor{blue}{0.114}$|$59.59 & \textcolor{blue}{0.091}$|$58.29 & \textcolor{blue}{0.077}$|$58.07 & \textcolor{blue}{0.071}$|$57.37 & \textcolor{blue}{0.065}$|$57.35 & \textcolor{blue}{0.149}$|$52.27 & \textcolor{blue}{0.131}$|$51.06 & 0.110$|$\textcolor{blue}{50.57} & \textcolor{blue}{0.094}$|$\textcolor{blue}{50.06} & 0.085$|$\textcolor{blue}{49.84} \\
\bottomrule % 底部横线
\end{tabular}
}
\vspace{-19pt}
\end{table*}
To evaluate the effectiveness of proposed batch and online frameworks, experiments are conducted on CIFAR100 \cite{Krizhevsky2009} and ImageNet \cite{Russakovsky2015} datasets, employing ResNet32 and ResNet50 \cite{He2016} as base classifiers. Training data follow standard long-tail imbalanced variants \cite{Liu2019}, while test priors employ both shuffled long-tail and Dirichlet-based imbalance settings to simulate realistic label shift. For both FMAPLS and online-FMAPLS, we initialize algorithms with a uniform prior ($\pi_{j}^{(0)}=1/K, \bm{\alpha}^{(0)}=\bm{1}$), which is fair and widely accepted due to the true test prior is unknown for the classifier. The performance of all methods is quantitatively assessed using the Kullback–Leibler (KL) divergence \cite{Csiszar1975} between the estimated and true label distributions:
\begin{align}
    \label{eq39}
    D_{\mathrm{KL}}(\bm{\pi}^{\text{true}}\|\bm{\pi}) = \sum_{j=1}^{K} \pi^{\text{true}}_j \log \frac{\pi^{\text{true}}_j}{\pi_j}
\end{align}
as well as the post-shift classification accuracy, where $\bm{\pi}^{\text{true}}$ is the true prior distribution of test dataset. Comparisons are made against several representative baselines, including MLLS \cite{Saerens2002}, BBSE \cite{Lipton2018}, RLLS \cite{Azizzadenesheli2019}, and MAPLS \cite{Ye2024}. All reported results are averaged over 100 independent Monte Carlo trials, each obtained through repeated random sampling of shuffled and Dirichlet-distributed test priors.

\begin{figure*}[t]
    \centering
    \vspace{-14pt}
    \subfloat[]{\includegraphics[width=1.76in]{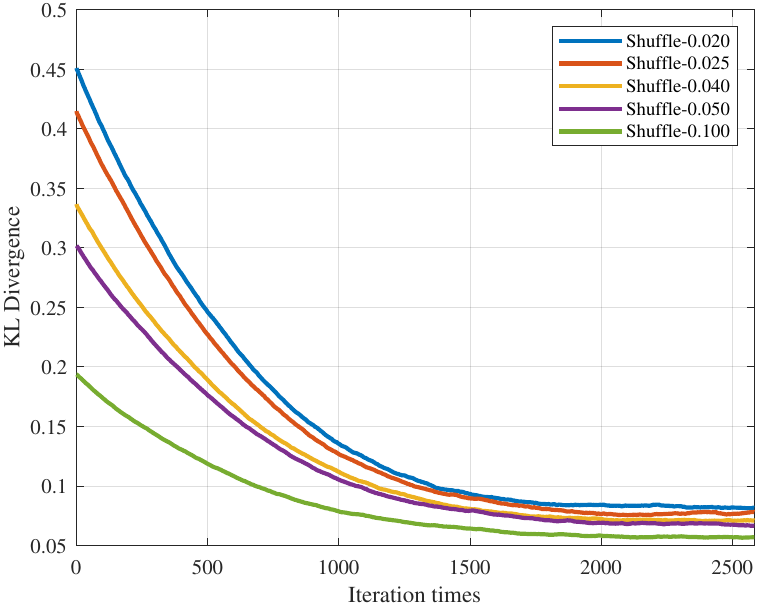}%
    \label{Fig2_first}}
    \hfil
    \subfloat[]{\includegraphics[width=1.76in]{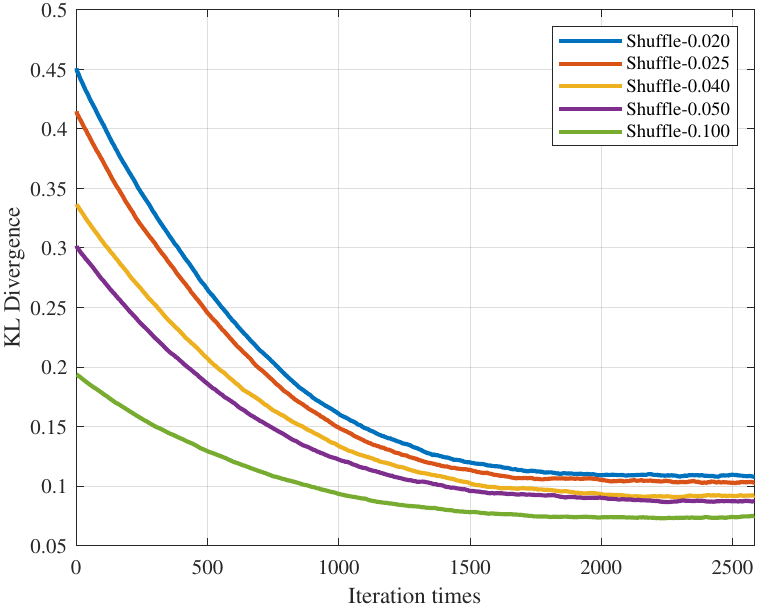}%
    \label{Fig2_second}}
    \hfil
    \subfloat[]{\includegraphics[width=1.76in]{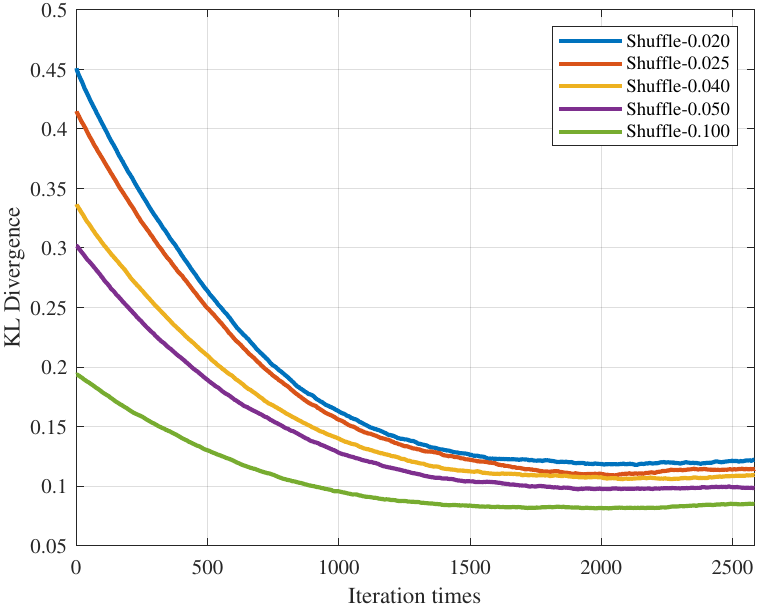}%
    \label{Fig2_thrid}}
    \hfil
    \subfloat[]{\includegraphics[width=1.76in]{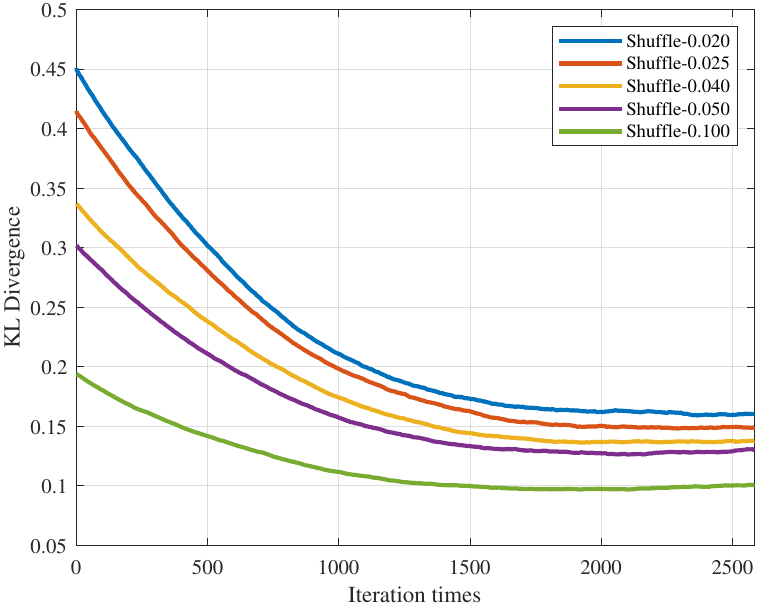}%
    \label{Fig2_fourth}}
    \vspace{-6pt}
    \caption{KL divergence of online-FMAPLS algorithm evaluated on CIFAR100 dataset with varying shuffled long-tail test imbalance ratio $\rho_{\text{test}}$ and different training imbalance ratios $\rho_{\text{train}}$. (a)~$\rho_{\text{train}}=5$, (b)~$\rho_{\text{train}}=10$, (c)~$\rho_{\text{train}}=20$, (d)~$\rho_{\text{train}}=50$.}
    \label{Figure2}
    \vspace{-20pt}
\end{figure*}

\begin{figure*}[t]
    \centering
    \subfloat[]{\includegraphics[width=1.76in]{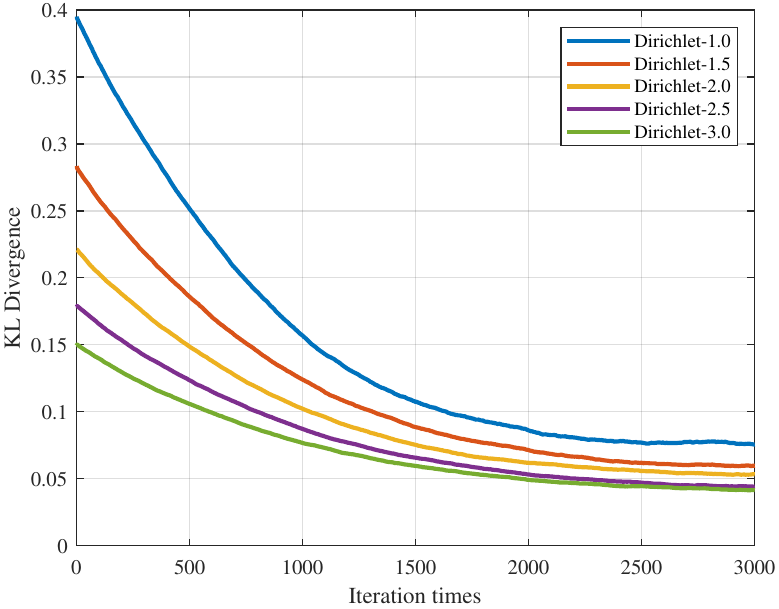}%
    \label{Fig3_first}}
    \hfil
    \subfloat[]{\includegraphics[width=1.76in]{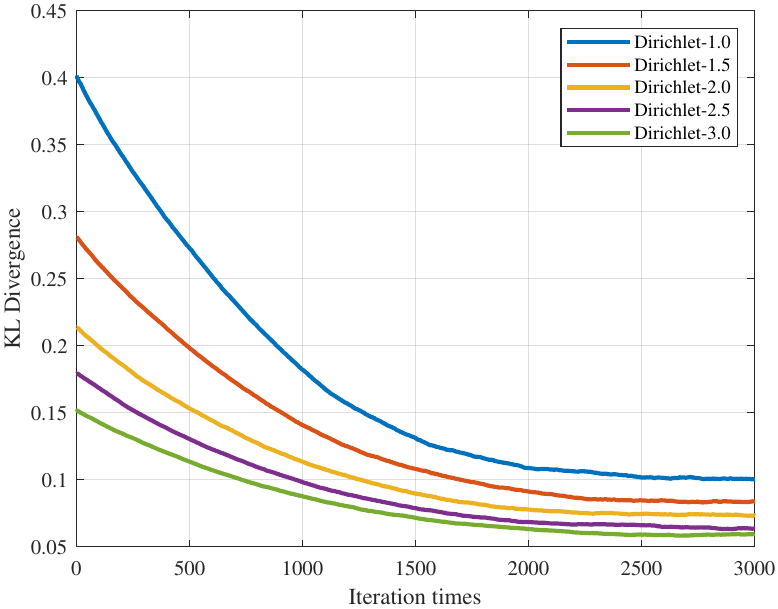}%
    \label{Fig3_second}}
    \hfil
    \subfloat[]{\includegraphics[width=1.76in]{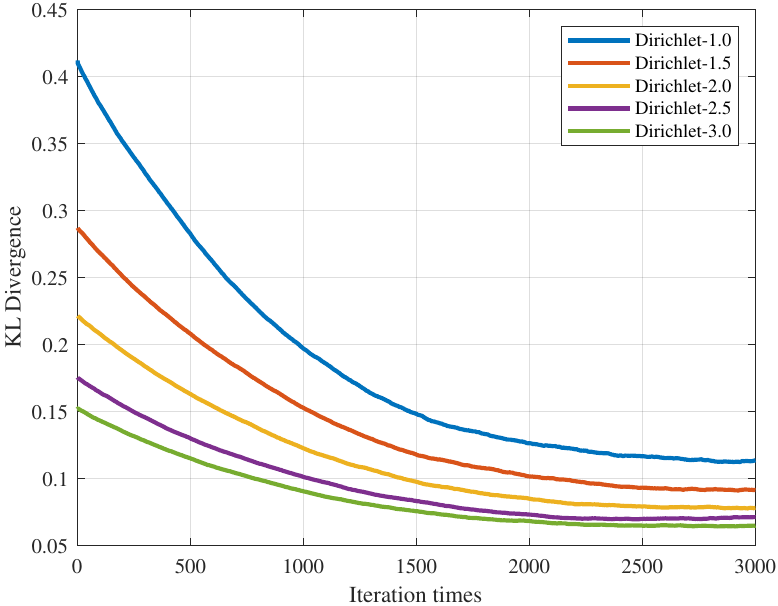}%
    \label{Fig3_thrid}}
    \hfil
    \subfloat[]{\includegraphics[width=1.76in]{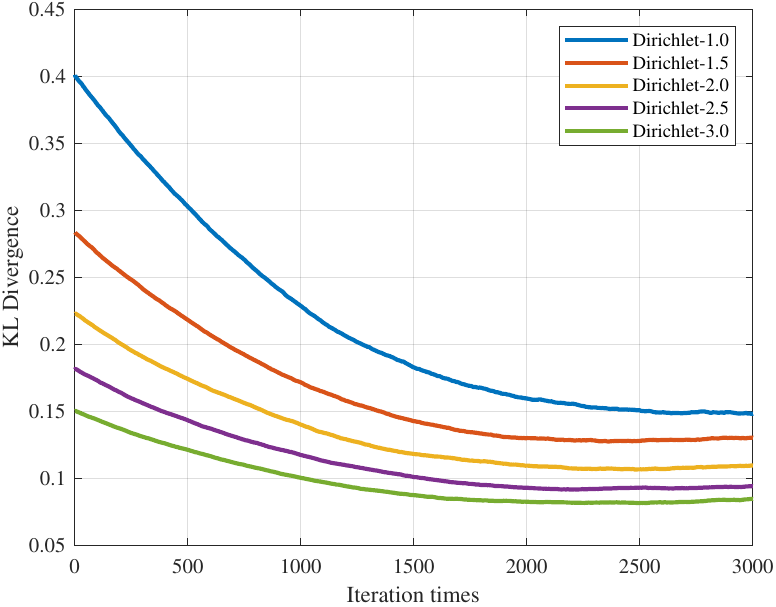}%
    \label{Fig3_fourth}}
    \vspace{-6pt}
    \caption{KL divergence of online-FMAPLS algorithm evaluated on CIFAR100 dataset with varying test prior Dirichlet hyperparameter $\bm{\alpha}$ and different training imbalance ratios $\rho_{\text{train}}$. (a)~$\rho_{\text{train}}=5$, (b)~$\rho_{\text{train}}=10$, (c)~$\rho_{\text{train}}=20$, (d)~$\rho_{\text{train}}=50$.}
    \label{Figure3}
    \vspace{-18pt}
\end{figure*}

\vspace{-12pt}
\subsection{Validation on CIFAR100} \label{Section6-1}
Table~\ref{Table2} demonstrates the label shift estimation KL divergence and post-shift classification accuracy of shuffled long-tailed test distributions, where the test imbalance ratio $\rho_{\text{test}}$ and training imbalance ratio $\rho_{\text{train}}$ are varied. 

From the perspective of KL divergence, the FMAPLS algorithm consistently achieves the lowest values across all parameter settings, with at least a 20\% reduction compared to baseline methods, demonstrating its superior effectiveness and robustness in estimating prior probability distributions for test data. In addition, both the training and test imbalance ratios significantly affect estimation precision: when $\rho_{\text{test}}$ is fixed, KL divergence increases as $\rho_{\text{train}}$ decreases, since a more imbalanced training set limits minority-class representation and thus degrades prior estimation; conversely, with fixed $\rho_{\text{train}}$, a smaller $\rho_{\text{test}}$ also raises KL due to the intensified label shift discrepancy from the uniform initialization. Nevertheless, the relative improvement of FMAPLS over baselines grows as $\rho_{\text{train}}$ decreases, indicating that the dynamic adaptation of the Dirichlet parameter $\bm{\alpha}$ enables more accurate target-prior approximation under severe imbalance. Moreover, under extremely imbalanced conditions ($\rho_{\text{train}}\leq0.05$), online-FMAPLS, which updates priors in a single pass, consistently attains the second-lowest KL, confirming its dynamic adaptive scheme remains effective in streaming scenarios.

From the perspective of classification accuracy, the proposed FMAPLS method consistently achieves top-tier performance, achieving the highest or second-highest accuracy across nearly all combinations of training and test imbalance ratios. This consistent superiority demonstrates FMAPLS's strong generalization capability under distribution shifts. While MLLS exhibits competitive, and occasionally superior, accuracy under severe test-time imbalance conditions (particularly when $\rho_{\text{test}}\leq0.025$), FMAPLS maintains superior overall robustness across the entire spectrum of test shifts. It preserves high accuracy not only in challenging training settings but also across most testing distributions, whereas the performance of MLLS declines more noticeably, and BBSE exhibits significant instability, as test conditions vary. Therefore, despite MLLS's situational advantage, FMAPLS establishes itself as a more universally reliable and high-performing approach for shuffled long-tailed label shift estimation.

Table~\ref{Table3} summarizes the KL divergence and classification accuracy of each algorithm under the Dirichlet test prior distribution. The experiments were conducted with the total number of test samples, $N_{\text{test}}$, fixed at 3000, and the Dirichlet hyperparameter $\bm{\alpha}_{\text{test}}$ set identically across all dimensions for simplicity. It is noteworthy that a smaller $\bm{\alpha}_{\text{test}}$ corresponds to higher uncertainty in the test prior distribution.

In terms of KL divergence, the proposed FMAPLS consistently achieves the lowest values across all training imbalance ratios and Dirichlet test hyperparameters $\bm{\alpha}_{\text{test}}$, while online-FMAPLS ranks second under highly imbalanced training conditions ($\rho_{\text{train}}\leq0.05$). This trend aligns with the shuffled long-tail results in Table~\ref{Table2}, confirming the effectiveness of dynamic Dirichlet parameter adaptation for test prior estimation. Moreover, as test uncertainty decreases (i.e., with larger $\bm{\alpha}_{\text{test}}$), all methods exhibit reduced KL divergence, indicating that more balanced and predictable test distributions ease label shift estimation. Notably, the performance gap between FMAPLS and baselines widens as $\bm{\alpha}_{\text{test}}$ decreases, highlighting that the proposed dynamic adaptation scheme is especially beneficial under challenging and uncertain test conditions.

In terms of classification accuracy, FMAPLS consistently achieves the best performance across almost all experimental conditions. Its robustness is particularly pronounced under severe training imbalance (e.g., 0.02-LT), where it significantly outperforms other methods. Furthermore, the performance gap between FMAPLS and other methods widens as the training imbalance intensifies. In contrast, the online-FMAPLS variant substantially reduces computational complexity while maintaining comparable accuracy to baseline algorithms, demonstrating its efficiency and scalability for large-scale and streaming data applications, albeit with a slight trade-off in predictive precision compared to the FMAPLS model.

To demonstrate the effectiveness of online-FMAPLS’s incremental prior updates, Fig.~\ref{Figure2} and Fig.~\ref{Figure3} depict the evolution of the KL divergence between the estimated and true test priors for the shuffled long-tailed and Dirichlet settings, respectively. At each iteration, the algorithm refines its prior estimates using the classifier’s soft outputs. By examining the curve profiles, the KL divergence decreases initially and then stabilizes as the iterations proceed, with convergence occurring around 1500 iterations for the shuffled long-tail and 2000 iterations for the Dirichlet distribution. These results confirm that online-FMAPLS achieves stable and accurate convergence through incremental updates.

\vspace{-12pt}
\subsection{Validation on ImageNet} \label{Section6-2}
\vspace{-3pt}
Table~\ref{Table4} and Table~\ref{Table5} present the KL divergence and classification accuracy on the large-scale long-tailed ImageNet dataset under shuffled and Dirichlet-distributed test priors. The proposed FMAPLS consistently achieves the lowest KL divergence across all settings, indicating its strong ability to estimate test priors accurately under highly imbalanced and uncertain conditions. The online-FMAPLS ranks second, highlighting the robustness of its incremental updates and the effectiveness of dynamic Dirichlet adaptation for large-scale streaming data. Moreover, the consistent improvements of FMAPLS and online-FMAPLS over all baselines demonstrate that relaxing restrictive assumptions in MAPLS enhances flexibility in modeling complex prior variations. Overall, these results validate the proposed frameworks’ scalability and generalization across diverse distributional scenarios.

Regarding the classification accuracy results in Tables~\ref{Table4} and~\ref{Table5}, FMAPLS consistently achieves the highest performance across nearly all configurations, indicating that precise estimation of test priors directly enhances post-shift performance. The online-FMAPLS attains comparable accuracy (only about 0.5\% lower in Table~\ref{Table4} and occasionally best in Table~\ref{Table5}) while requiring far less computation, underscoring its practicality for large-scale and streaming scenarios. Furthermore, both methods maintain stable accuracy as $\rho_{\text{test}}$ or $\bm{\alpha}_{\text{test}}$ increases, confirming robustness to varying degrees of imbalance and uncertainty. Interestingly, most methods exhibit a slight accuracy drop when $\rho_{\text{test}}$ or $\bm{\alpha}_{\text{test}}$ grows. This phenomenon arises because the classifier, trained on a long-tailed source, tends to align its decision boundaries with skewed class priors; when the test prior becomes more uniform and less uncertain, large calibration mismatch leads to a marginal degradation in overall accuracy.

\begin{table}[b]
\vspace{-23pt}
\centering
\caption{KL Divergence and Classification Accuracy (“KL$|$Acc.”) on Shuffled Long-Tail Test Priors with Test Ratios $\{0.02, 0.025, 0.04, 0.05\}$. Best Performances are in Bold and Second Best in Blue.}
\vspace{-7pt}
\label{Table4}
{\scriptsize
\setlength{\tabcolsep}{3.5pt}
\begin{tabular}{c|cccc} % 指定表格列数和对齐方式
\toprule % 顶部横线
Dataset & \multicolumn{4}{c}{ImageNet-LT}\\ % 表头
\midrule % 中间横线
Imbalance ratio $\rho_{\text{test}}$ & 0.02 & 0.025 & 0.04 & 0.05 \\
\midrule % 中间横线
MLLS & 1.584$|$46.02 & 1.494$|$46.08 & 1.582$|$44.82 & 1.545$|$44.43 \\
BBSE & 0.351$|$50.93 & 0.340$|$50.59 & 0.313$|$50.17 & 0.300$|$50.18 \\
RLLS & 0.967$|$45.81 & 0.936$|$45.75 & 0.865$|$45.79 & 0.818$|$46.08 \\
MAPLS & 0.229$|$51.20 & 0.220$|$51.20 & 0.197$|$50.28 & 0.193$|$50.07 \\
\midrule % 中间横线
\textbf{FMAPLS} & \textbf{0.180}$|$\textbf{52.78} & \textbf{0.171}$|$\textbf{52.49} & \textbf{0.148}$|$\textbf{51.77} & \textbf{0.142}$|$\textbf{51.68} \\
\textbf{online-FMAPLS} & \textcolor{blue}{0.224}$|$\textcolor{blue}{52.22} & \textcolor{blue}{0.213}$|$\textcolor{blue}{51.95} & \textcolor{blue}{0.187}$|$\textcolor{blue}{51.22} & \textcolor{blue}{0.175}$|$\textcolor{blue}{51.21} \\
\bottomrule % 底部横线
\end{tabular}
}
\vspace{-14pt}
\end{table}
\begin{table}[b]
\centering
\caption{KL Divergence and Classification Accuracy (“KL$|$Acc.”) on Dirichlet Test Priors with Hyperparameters $\{1, 1.5, 2, 2.5\}$. Best Performances are in Bold and Second Best in Blue.}
\vspace{-7pt}
\label{Table5}
{\scriptsize
\setlength{\tabcolsep}{3.5pt}
\begin{tabular}{c|cccc} % 指定表格列数和对齐方式
\toprule % 顶部横线
Dataset & \multicolumn{4}{c}{ImageNet-LT} \\
\midrule % 中间横线
Dirichlet $\bm{\alpha}_{\text{test}}$ & 1 & 1.5 & 2 & 2.5 \\
\midrule % 中间横线
MLLS & 1.504$|$45.03 & 1.554$|$44.06 & 1.534$|$43.49 & 1.595$|$42.49 \\
BBSE & 0.335$|$50.73 & 0.299$|$50.06 & 0.281$|$49.51 & 0.267$|$49.28 \\
RLLS & 0.935$|$45.81 & 0.812$|$45.98 & 0.750$|$45.89 & 0.712$|$45.92 \\
MAPLS & 0.227$|$50.41 & 0.191$|$49.90 & 0.163$|$49.64 & 0.164$|$48.95 \\
\midrule % 中间横线
\textbf{FMAPLS} & \textbf{0.173}$|$\textbf{52.07} & \textbf{0.143}$|$\textbf{51.52} & \textbf{0.123}$|$\textcolor{blue}{50.89} & \textbf{0.109}$|$\textcolor{blue}{50.76} \\
\textbf{online-FMAPLS} & \textcolor{blue}{0.202}$|$\textcolor{blue}{51.90} & \textcolor{blue}{0.159}$|$\textcolor{blue}{51.51} & \textcolor{blue}{0.133}$|$\textbf{51.04} & \textcolor{blue}{0.113}$|$\textbf{50.86} \\
\bottomrule % 底部横线
\end{tabular}
}
\vspace{-9pt}
\end{table}

\begin{figure}[t]
    \vspace{-12pt}
    \centering
    \subfloat[] {
        \includegraphics[width=1.68in]{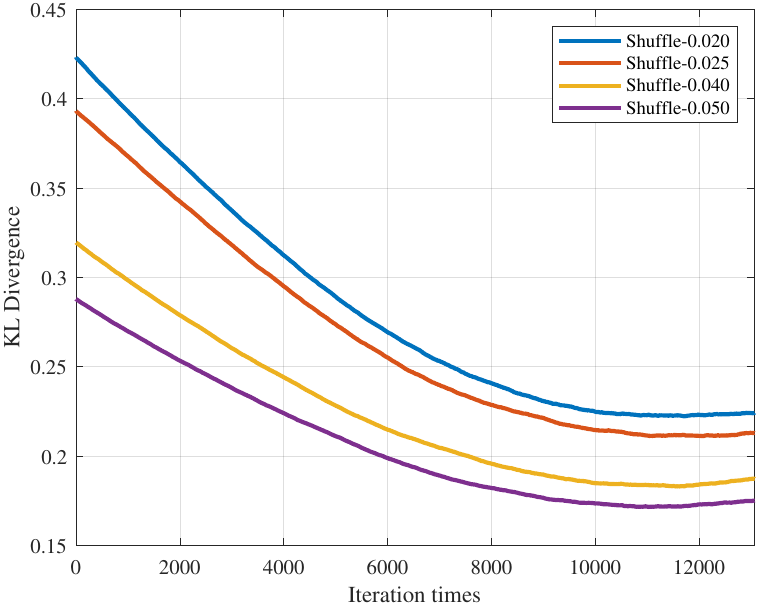}
        \label{Figure4-1}
    }
    \subfloat[] {
        \includegraphics[width=1.69in]{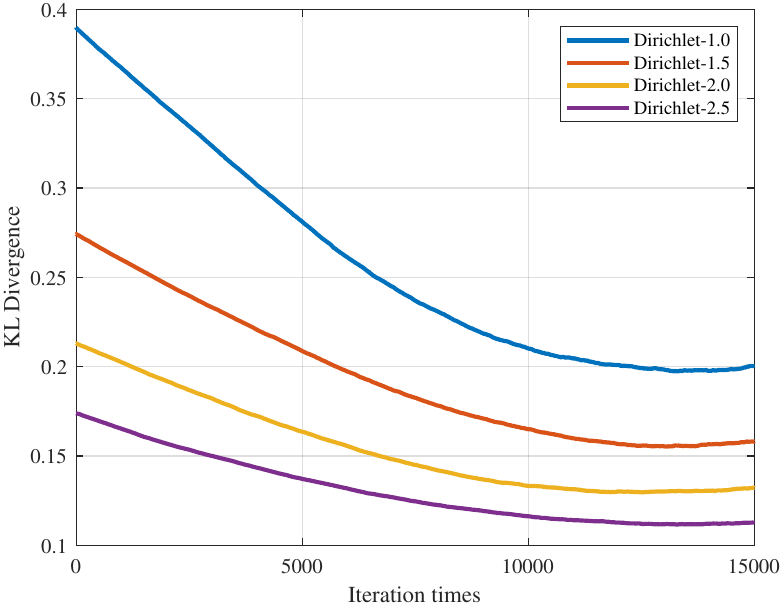}
        \label{Figure4-2}
    }
    \vspace{-5pt}
    \caption{KL divergence of online-FMAPLS algorithm evaluated on long-tail ImageNet dataset with (a)~varying shuffled long-tail test imbalance ratio $\rho_{\text{test}}$, (b)~varying test Dirichlet hyperparameter $\bm{\alpha}$.}
    \label{Figure4}
    \vspace{-20pt}
\end{figure}

Similar to Figs.~\ref{Figure2} and~\ref{Figure3}, Figs.~\ref{Figure4}(a) and~\ref{Figure4}(b) depict the evolution of KL divergence between the estimated and true test priors for online-FMAPLS on the large-scale ImageNet-LT dataset. In the shuffled long-tailed setting, the KL divergence gradually decreases and stabilizes after about 10000 iterations, while in the Dirichlet prior case, convergence occurs around 11000 iterations. Although the convergence rate is slower compared with CIFAR100 dataset, primarily due to the increased data volume and the much larger number of classes, the online algorithm still exhibits clear convergence behavior. These results confirm that online-FMAPLS effectively handles high-dimensional streaming data and accurately estimates complex label-shifted priors in large-scale recognition tasks.

\vspace{-12pt}
\subsection{Ablation comparison and trade-off analysis} \label{Section6-3}
In this subsection, we perform an ablation study comparing three Bayesian-based variants to examine the impact of model assumptions and algorithmic design choices. The baseline MAPLS algorithm employs fixed Dirichlet hyperparameters $\boldsymbol{\alpha}$, imposing rigid prior constraints. FMAPLS relaxes these constraints and enables dynamic adaptation of $\boldsymbol{\alpha}$ for more flexible prior modeling. Building upon FMAPLS, the online-FMAPLS algorithm further extends this framework to support streaming data processing through incremental updates. It is worth noting that the ablation analysis focuses solely on the KL divergence results under the most challenging experimental settings, namely the most imbalanced test prior ($\rho_{\text{test}} = 0.02$) and the highest uncertainty case ($\boldsymbol{\alpha}_{\text{test}} = \bm{1}$).

\begin{figure}[!t]
    \centering
    \subfloat[] {
        \includegraphics[width=2.5in]{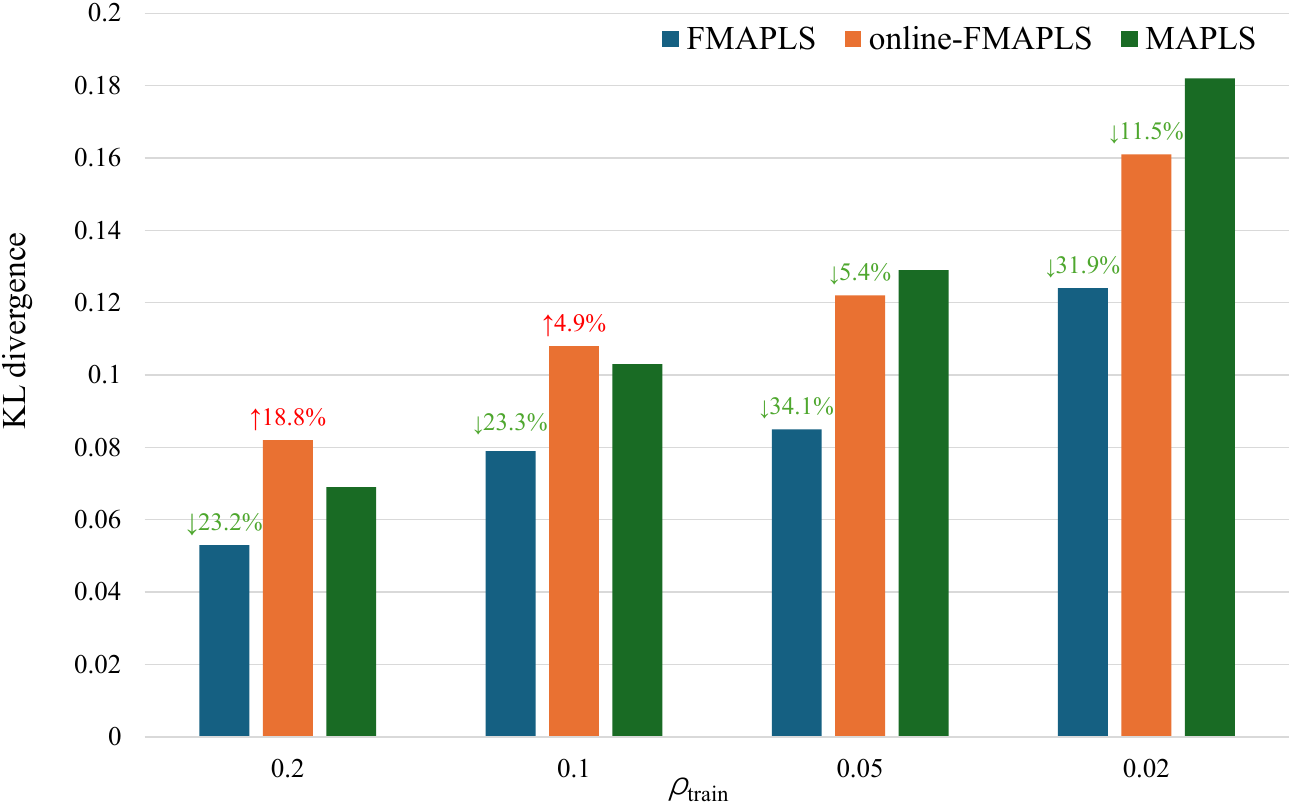}
        \label{Figure5-1}
    }
    \vspace{-1pt}
    \subfloat[] {
        \includegraphics[width=2.5in]{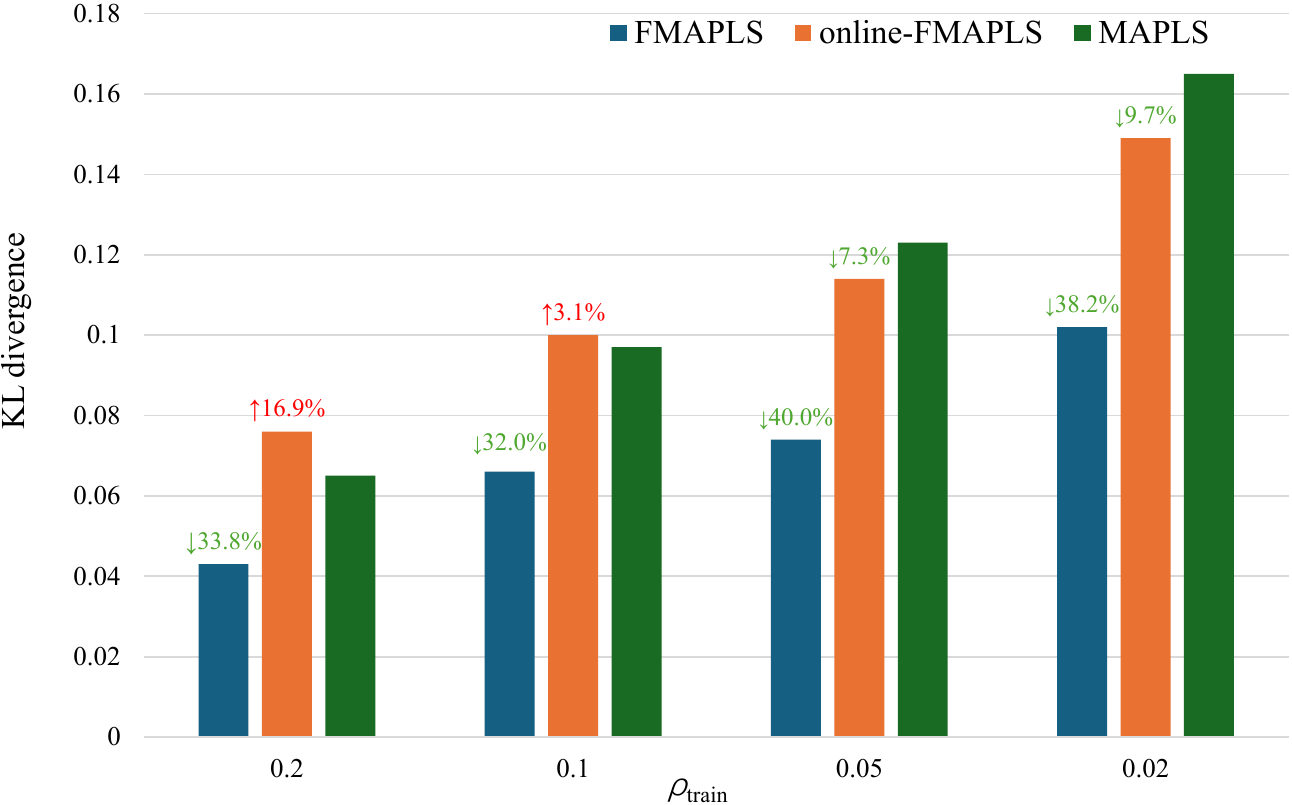}
        \label{Figure5-2}
    }
    \vspace{-2pt}
    \caption{Ablation comparison of FMAPLS, online FMAPLS, and MAPLS under different training prior distributions. (a)~Shuffled long-tail imbalance ratio $\rho_{\text{test}}=0.02$, (b)~Test Dirichlet hyperparameter $\bm{\alpha}_{\text{test}}=\bm{1}$.}
    \label{Figure5}
    \vspace{-6pt}
\end{figure}

Fig.~\ref{Figure5} illustrates the KL divergence performance of the above three algorithms on the CIFAR100 dataset. When the training imbalance ratio is relatively large ($\rho_{\text{train}} \geq 0.1$), the proposed FMAPLS reduces KL divergence by about 23\% and 33\% under the shuffled and Dirichlet test settings, respectively, compared with MAPLS. However, the KL performance of online-FMAPLS algorithm is slightly inferior to that of MAPLS in both test environments. This degradation arises due to the online version relies on single-pass updates, which limit the utilization of global statistical information and thereby increase estimation variance compared with batch-based methods. When $\rho_{\text{train}}<0.1$, the proposed FMAPLS algorithm achieves reductions of about 33\% and 40\% in KL divergence under above two environments, while the final convergence performance of the online-FMAPLS algorithm also surpasses that of MAPLS. These results further confirm the effectiveness of dynamic parameter adaptation in enhancing prior estimation accuracy under severely imbalanced conditions.

\begin{figure}[!t]
    \centering
    \includegraphics[width=2.5in]{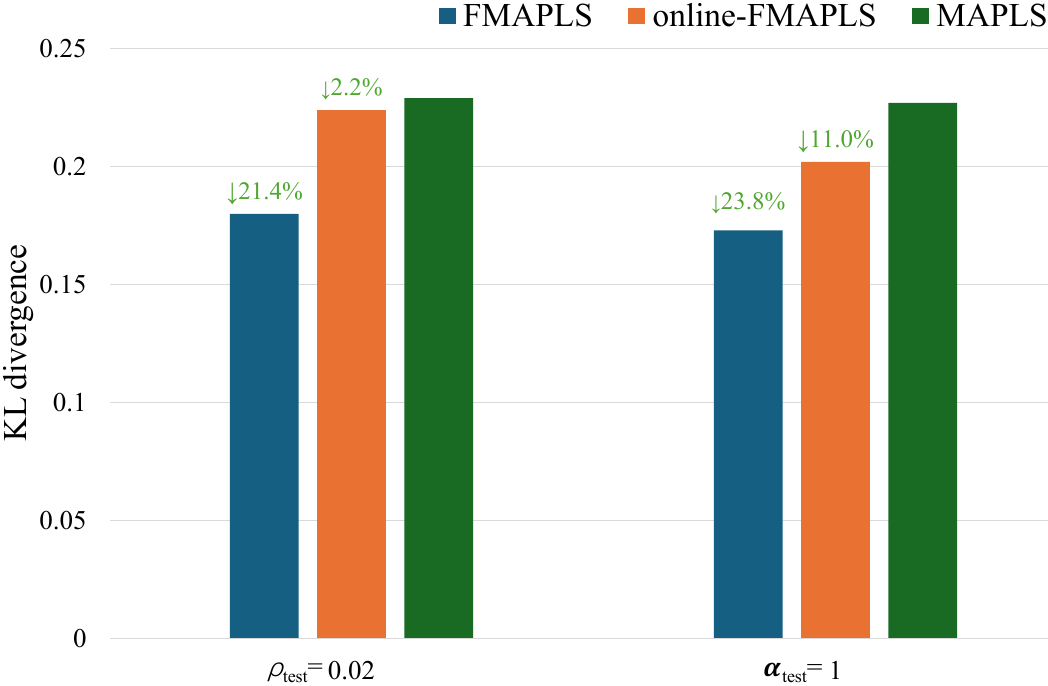}
    \vspace{-6pt}
    \caption{Ablation comparison of FMAPLS, online FMAPLS, and MAPLS under long-tail ImageNet training dataset.}
    \label{Figure6}
    \vspace{-16pt}
\end{figure}

Fig.~\ref{Figure6} presents the ablation comparison results on the long-tailed ImageNet dataset. The experimental data indicate that, regardless of whether the test scenario involves extreme imbalance or high uncertainty, both FMAPLS and its online variant outperform the batch-based MAPLS method. Specifically, when $\rho_{\text{test}} = 0.02$, FMAPLS achieves a 21.4\% reduction in KL divergence compared with MAPLS, while the online-FMAPLS maintains comparable performance with a 2.2\% degradation. Under the Dirichlet prior with $\bm{\alpha}_{\text{test}} = \bm{1}$, FMAPLS attains a 23.8\% improvement and the online version exhibits an 11.0\% decrease in KL divergence relative to MAPLS. These results confirm that the proposed dynamic parameter adaptation mechanism effectively enhances estimation robustness and maintains stable performance under ImageNet dataset.

\section{Conclusion} \label{Section7}
In this paper, we developed a Bayesian-based framework for label shift estimation that integrates dynamic hyperparameter adaptation and efficient online learning mechanisms. The proposed FMAPLS algorithm extends the conventional MAPLS approach by jointly optimizing Dirichlet hyperparameters $\bm{\alpha}$ and class priors $\bm{\pi}$, thereby enhancing flexibility in modeling complex test distributions. To further improve scalability, the LSF method is introduced to provide closed-form Dirichlet hyperparameter updates, which effectively reducing computational complexity. The online-FMAPLS algorithm further extends this framework to streaming environments by incorporating stochastic approximation, thereby enabling real-time and efficient adaptation to sequentially arriving data. Furthermore, we analyze and demonstrate a fundamental trade-off between the online convergence rate and estimation accuracy. Extensive experiments and ablation comparisons on CIFAR100 and ImageNet datasets confirm that the proposed methods achieve superior estimation KL divergence and post-shift classification accuracy across various label shift scenarios, maintaining robustness and stability even under severe imbalance and uncertain test distributions.

\vspace{-10pt}
{\appendices 
\section{Proof of equation (\ref{eq7})} \label{Appendix-1}
Substitute the PDF of the Dirichlet distribution $Dir(\bm{\pi}; \bm{\alpha})$ into equation (\ref{eq6}), the complete log probability can be expressed as:
\begin{align}
    \label{A-1-1}
    \log P(\boldsymbol{\theta}&|\boldsymbol{X}, \boldsymbol{Y})=\log \frac{1}{Z}+\log\frac{1}{\boldsymbol{B}(\boldsymbol{\alpha})}\prod_{j=1}^{K}\pi_{j}^{\alpha_{j}-1} \\
    &+\log\prod_{i=1}^{N}P(X_s=x_i|Y_s=y_i)P(Y_t=y_i|\boldsymbol{\theta}) \nonumber \\
    &=Const+\log\frac{1}{\boldsymbol{B}(\boldsymbol{\alpha})}+\sum_{j=1}^{K}(\alpha_j-1)\log\pi_j \nonumber \\
    &+\sum_{i=1}^{N}\log P(Y_t=y_i|\boldsymbol{\theta}) \nonumber \\
    &=Const+\log\frac{1}{\boldsymbol{B}(\boldsymbol{\alpha})}+\sum_{j=1}^{K}(\alpha_j-1)\log\pi_j \nonumber \\
    &+\sum_{i=1}^{N}\log\prod_{j=1}^{K}\pi_j^{\delta_{y_i,j}} \nonumber \\
    &=Const+\log\frac{1}{\boldsymbol{B}(\boldsymbol{\alpha})}+\sum_{j=1}^{K}(\alpha_j-1)\log\pi_j \nonumber \\
    &+\sum_{i=1}^{N}\sum_{j=1}^{K}\delta_{y_i,j}\log\pi_j \nonumber
\end{align}
where $\delta_{y_i,j}$ is the Kronecker delta function. In the E step, given parameter $\boldsymbol{\theta}^{(t)}$ at $t$-th iteration, the expectation of complete-data logarithm function can be written as:
\begin{align}
    \label{A-1-2}
    &\mathbb{E}_{\boldsymbol{Y}|\boldsymbol{X}, \boldsymbol{\theta}^{(t)}} \lbrack \log P(\boldsymbol{\theta}|\boldsymbol{X}, \boldsymbol{Y})\rbrack \\
    % &=\mathbb{E}_{\boldsymbol{Y}|\boldsymbol{X}, \boldsymbol{\theta}^{(t)}}\Big\lbrack Const+\log\frac{1}{\boldsymbol{B}(\boldsymbol{\alpha})} \nonumber \\
    % &+\sum_{j=1}^{K}(\alpha_j-1)\log\pi_j+\sum_{i=1}^{N}\sum_{j=1}^{K}\delta_{y_i,j}\log\pi_j\Big\rbrack \nonumber \\
    &=Const+\log\frac{1}{\boldsymbol{B}(\boldsymbol{\alpha})}+\sum_{j=1}^{K}(\alpha_j-1)\log\pi_j \nonumber \\
    &+\sum_{i=1}^{N}\sum_{j=1}^{K}E_{\boldsymbol{Y}|\boldsymbol{X}, \boldsymbol{\theta}^{(t)}}\Big\lbrack \delta_{y_i,j}\log\pi_j\Big\rbrack \nonumber \\
    &=Const+\log\frac{1}{\boldsymbol{B}(\boldsymbol{\alpha})}+\sum_{j=1}^{K}(\alpha_j-1)\log\pi_j \nonumber \\
    &+\sum_{j=1}^{K}\sum_{i=1}^{N}P(Y_t=y_i=j|X_t=x_i,\boldsymbol{\theta}^{(t)})\log\pi_j \nonumber
\end{align}

\vspace{-5pt}
\section{Proof of the FIM elements estimated by \\ FMAPLS algorithm for $\bm{\pi}$} \label{Appendix-2}
Rewrite the $\bm{\pi}$ log-likelihood function with normalization constraint:
\begin{align}
    \label{A-2-1}
    \left\{
        \begin{aligned}
        &\underbrace{\sum_{j=1}^{K}(\alpha_j^{(t)}-1)\log\pi_j+\sum_{j=1}^{K}\sum_{n=1}^{N}P(y_n=j|x_n,\bm{\theta}^{(t)})\log\pi_j}_{L_{\pi}}\\
        &\text{s.t.}: \;\; \sum_{j=1}^{K}\pi_j=1 \;\; \text{and} \;\; \pi_j>0\\
    \end{aligned}
    \right.
\end{align}
Due to the prior probability $\bm{\pi}$ is distributed in a $(K-1)$-dimensional standard simplex, $L_{\pi}$ can further be written as:
\begin{align}
    \label{A-2-2}
    L_{\pi}=L_{\pi_{k}}+L_{\pi_{K}} \;\;\;\;\;1\leq k\leq K-1
\end{align}
\vspace{-20pt}
\begin{align}
    \label{A-2-3}
    \pi_{K}=1-\sum_{k=1}^{K-1}\pi_k
\end{align}
Thus, the elements of $L_{\pi}$ Hessian matrix can be expressed as:
\begin{align}
    \label{A-2-4}
    \frac{\partial L_{\pi}}{\partial \pi_i}&=\frac{\partial L_{\pi_{k}}}{\partial \pi_i}+\frac{\partial L_{\pi_{K}}}{\partial \pi_i} \\
    &=\frac{m_i}{\pi_i}+\frac{\partial \left(m_K\log(1-\sum_{k=1}^{K}\pi_k)\right)}{\partial \pi_i} \nonumber \\
    &=\frac{m_i}{\pi_i}-\frac{m_K}{\pi_K} \nonumber
\end{align}
\vspace{-15pt}
\begin{align}
    \label{A-2-5}
    \frac{\partial^2 L_{\pi}}{\partial \pi_i \partial \pi_j}&=-\frac{m_i}{\pi_i^2}\delta_{i,j}-\frac{\partial}{\partial \pi_j}(\frac{m_K}{\pi_K}) \\
    &=-\frac{m_i}{\pi_i^2}\delta_{i,j}-\frac{\partial}{\partial \pi_K}(\frac{m_K}{\pi_K})\frac{\partial \pi_K}{\partial \pi_j} \nonumber \\
    &=-\frac{m_i}{\pi_i^2}\delta_{i,j}-\frac{m_K}{\pi_K^2} \;\;\;\;\;\;\;\; (K-1)\times (K-1) \nonumber
\end{align}
where $m_i=\alpha^{(t)}_i-1+\sum_{n=1}^NP(y_n=i|x_n,\bm{\theta}^{(t)})$. On the basis of Hessian matrix, the elements of FIM can be given as:
\begin{align}
    \label{A-2-6}
    F_{i,j}&=-\mathbb{E}\Big[\frac{\partial^2 L_{\pi}}{\partial \pi_i \partial \pi_j}\Big] \\
    &=\mathbb{E}\Big[\frac{\alpha^{(t)}_i-1+\sum_{n=1}^NP(y_n=i|x_n,\bm{\theta}^{(t)})}{\pi_i^2}\delta_{i,j} \nonumber \\
    &+\frac{\alpha^{(t)}_K-1+\sum_{n=1}^NP(y_n=K|x_n,\bm{\theta}^{(t)})}{\pi_K^2} \Big] \nonumber \\
    &=\frac{\alpha^{(t)}_i-1+\sum_{n=1}^N\mathbb{E}_{x_n}[P(y_n=i|x_n,\bm{\theta}^{(t)})]}{\pi_i^2}\delta_{i,j} \nonumber \\
    &+\frac{\alpha^{(t)}_K-1+\sum_{n=1}^N\mathbb{E}_{x_n}[P(y_n=K|x_n,\bm{\theta}^{(t)})]}{\pi_K^2}  \nonumber \\
    &=\frac{\alpha^{(t)}_i-1+N\mathbb{E}_X[f_i^{X}]}{\pi_i^2}\delta_{i,j}+\frac{\alpha^{(t)}_K-1+\mathbb{E}_X[f_K^{X}]}{\pi_K^2} \nonumber
\end{align}
where $f_i^X$ denotes the posterior probability that the classifier assigns the input sample $X$ to class $i$. However, in practical scenarios, the expectation $\mathbb{E}_X[f_i^{X}]$ is typically unavailable; therefore, we use the sample mean to approximate it:
\begin{align}
    \label{A-2-7}
    \mathbb{E}_X[f_i^{X}]\approx\frac{1}{N}\sum_{n=1}^Nf_i^{x_n}
\end{align}
Substituting (\ref{A-2-7}) into (\ref{A-2-6}) leads to the equation (\ref{eq14}).

\vspace{-8pt}
\section{Proof of (\ref{eq21}) belongs to exponential family} \label{Appendix-3}
Substituting the PDF of the Dirichlet distribution $Dir(\bm{\pi}; \bm{\alpha})$ into equation (\ref{eq21}) yields the following expression:
\begin{align}
    \label{A-3-1}
    &P(\boldsymbol{\theta}|X, Y) \\
    &=e^{\log\left(\frac{1}{Z}\frac{1}{\boldsymbol{B}(\boldsymbol{\alpha})}\prod_{i=1}^{K}\pi_{i}^{\alpha_{i}-1}P(X_s=x^\tau|Y_s=y^\tau)\prod_{j=1}^{K}\pi_{j}^{\delta_{y^\tau,j}}\right)} \nonumber \\
    &=e^{\log\left(\frac{1}{Z}P(X_s=x^\tau|Y_s=y^\tau)\right)+\log\left(\frac{1}{\boldsymbol{B}(\boldsymbol{\alpha})}\prod_{i=1}^{K}\pi_{i}^{\alpha_{i}-1}\prod_{j=1}^{K}\pi_{j}^{\delta_{y^\tau,j}}\right)} \nonumber \\
    &=Ae^{\log\left(\frac{1}{\boldsymbol{B}(\boldsymbol{\alpha})}\right)+\sum_{i=1}^K(\alpha_i-1+\delta_{y^\tau,i})\log\pi_i} \nonumber \\
    &=Ae^{\log\left(\frac{1}{\boldsymbol{B}(\boldsymbol{\alpha})}\right)+\left\langle \begin{pmatrix}
                  1 \\
                  \delta_{y^\tau,i}
              \end{pmatrix}_i , \begin{pmatrix}
                  (\alpha_i-1)\log\pi_i \\
                  \log\pi_i
              \end{pmatrix}_i\right\rangle} \nonumber
\end{align}
\vspace{-12pt}
\begin{align}
    \label{A-3-2}
    A = e^{\log\left(\frac{1}{Z}P(X_s=x^\tau|Y_s=y^\tau)\right)}
\end{align}
Based on the derivation of (\ref{A-3-1}), it is easy to see that the complete posterior takes the form of equation (\ref{eq21}).

\vspace{-8pt}
\section{Proof of the FIM elements estimated by \\ online-FMAPLS algorithm for $\bm{\pi}$} \label{Appendix-4}
According to the online algorithm's $\bm{\pi}$ log-likelihood function (\ref{eq26}) and the derive in Appendix~\ref{Appendix-2}, the elements of Hessian matrix can be expressed as:
\begin{align}
    \label{A-4-1}
    \frac{\partial^2 L_{\pi}}{\partial \pi_i \partial \pi_j}&=-\frac{n_i}{\pi_i^2}\delta_{i,j}-\frac{\partial}{\partial \pi_j}(\frac{n_K}{\pi_K}) \\
    &=-\frac{n_i}{\pi_i^2}\delta_{i,j}-\frac{n_K}{\pi_K^2} \;\;\;\;\;\;\;\; (K-1)\times (K-1) \nonumber
\end{align}
where $n_i=\alpha_i^{\tau}-1+(1-\gamma)\delta_{y^{\tau},i}+\gamma P(y^{\tau+1}=i|x^{\tau+1},\bm{\theta}^{\tau})$. On the basis of Hessian matrix, the elements of Fisher information matrix can be further written as:
\begin{align}
    \label{A-4-2}
    &F_{i,j}=-\mathbb{E}\left[\frac{\partial^2 L_{\pi}}{\partial \pi_i \partial \pi_j}\right] \\
    % &=\mathbb{E}\big[\frac{\alpha_i^{\tau}-1+(1-\gamma)\delta_{y^{\tau},i}+\gamma P(y^{\tau+1}=i|x^{\tau+1},\bm{\theta}^{\tau})}{\pi_i^2}\delta_{i,j} \nonumber \\
    % &+\frac{\alpha_K^{\tau}-1+(1-\gamma)\delta_{y^{\tau},K}+\gamma P(y^{\tau+1}=K|x^{\tau+1},\bm{\theta}^{\tau})}{\pi_K^2} \big] \nonumber \\
    &=\frac{\alpha_i^{\tau}-1+(1-\gamma)\delta_{y^{\tau},i}+\gamma\mathbb{E}_X[f_i^{X}]}{\pi_i^2}\delta_{i,j} \nonumber \\
    &+\frac{\alpha_K^{\tau}-1+(1-\gamma)\delta_{y^{\tau},K}+\gamma\mathbb{E}_X[f_K^{X}]}{\pi_K^2} \nonumber
\end{align}
Using the sample mean to replace $\mathbb{E}_X[f_i^{X}]$ leads to (\ref{eq34}):
\begin{align}
    \label{A-4-3}
    \mathbb{E}_X[f_i^{X}]\approx\frac{1}{\tau}\sum_{t=1}^\tau f_i^{x^t}
\end{align}
}

% \bf{If you include a photo:}\vspace{-33pt}
% \begin{IEEEbiography}[{\includegraphics[width=1in,height=1.25in,clip,keepaspectratio]{fig1}}]{Michael Shell}
% Use $\backslash${\tt{begin\{IEEEbiography\}}} and then for the 1st argument use $\backslash${\tt{includegraphics}} to declare and link the author photo.
% Use the author name as the 3rd argument followed by the biography text.
% \end{IEEEbiography}

% \vspace{11pt}

% \bf{If you will not include a photo:}\vspace{-33pt}
% \begin{IEEEbiographynophoto}{John Doe}
% Use $\backslash${\tt{begin\{IEEEbiographynophoto\}}} and the author name as the argument followed by the biography text.
% \end{IEEEbiographynophoto}
\vfill

\end{document}